\def\year{2022}\relax
\documentclass[letterpaper]{article} 
\usepackage{aaai22}  
\usepackage{times}  
\usepackage{helvet}  
\usepackage{courier}  
\usepackage[hyphens]{url}  
\usepackage{graphicx, color} 
\usepackage{natbib}  
\usepackage{caption} 
\usepackage{amssymb}
\usepackage{amsmath}
\usepackage{multirow}
\usepackage{multicol}
\usepackage{booktabs}
\usepackage{threeparttable}
\usepackage{bbding}
\usepackage{verbatim}
\usepackage{float}
\usepackage{stfloats}
\DeclareCaptionStyle{ruled}{labelfont=normalfont,labelsep=colon,strut=off} 
\usepackage{algorithm}
\usepackage{algorithmic}

\usepackage{newfloat}
\usepackage{listings}
\newcommand{\tabincell}[2]{\begin{tabular}{@{}#1@{}}#2\end{tabular}} 
\floatstyle{ruled}
\newfloat{listing}{tb}{lst}{}
\floatname{listing}{Listing}
\title{Proximal PanNet: A Model-Based Deep Network for Pansharpening}
\author {
    Xiangyong Cao\textsuperscript{\rm 1,\rm 2},
    Yang Chen\textsuperscript{\rm 3},
    Wenfei Cao\textsuperscript{\rm 3$\star$}
}
\affiliations {
    \textsuperscript{\rm 1} School of Electronic and Information Engineering, Xi’an Jiaotong University, Xi’an 710049, China.\\
    \textsuperscript{\rm 2} Ministry of Education Key Lab For Intelligent Networks and Network Security, Xi’an 710049, China.\\
    \textsuperscript{\rm 3}  School of Mathematics and Statistics, Shaanxi Normal University, Xi’an 710119, China.\\
    caoxiangyong@mail.xjtu.edu.cn, yangchen9103@163.com, caowenf2015@snnu.edu.cn
}

\usepackage{bibentry}

\begin{document}

\maketitle

\begin{abstract}
Recently, deep learning techniques have been extensively studied for pansharpening, which aims to generate a high resolution multispectral (HRMS) image by fusing a low resolution multispectral (LRMS) image with a high resolution panchromatic (PAN) image. However, existing deep learning-based pansharpening methods directly learn the mapping from LRMS and PAN to HRMS. These network architectures always lack sufficient interpretability, which limits further performance improvements. To alleviate this issue, we propose a novel deep network for pansharpening by combining the model-based methodology with the deep learning method. Firstly, we build an observation model for pansharpening using the convolutional sparse coding (CSC) technique and design a proximal gradient algorithm to solve this model. Secondly, we unfold the iterative algorithm into a deep network, dubbed as \textit{Proximal PanNet}, by learning the proximal operators using convolutional neural networks. Finally, all the learnable modules can be automatically learned in an end-to-end manner. Experimental results on some benchmark datasets show that our network performs better than other advanced methods both quantitatively and qualitatively.
\end{abstract}

\section{Introduction}
\begin{figure}
	\centering
	\includegraphics[width=0.9\linewidth]{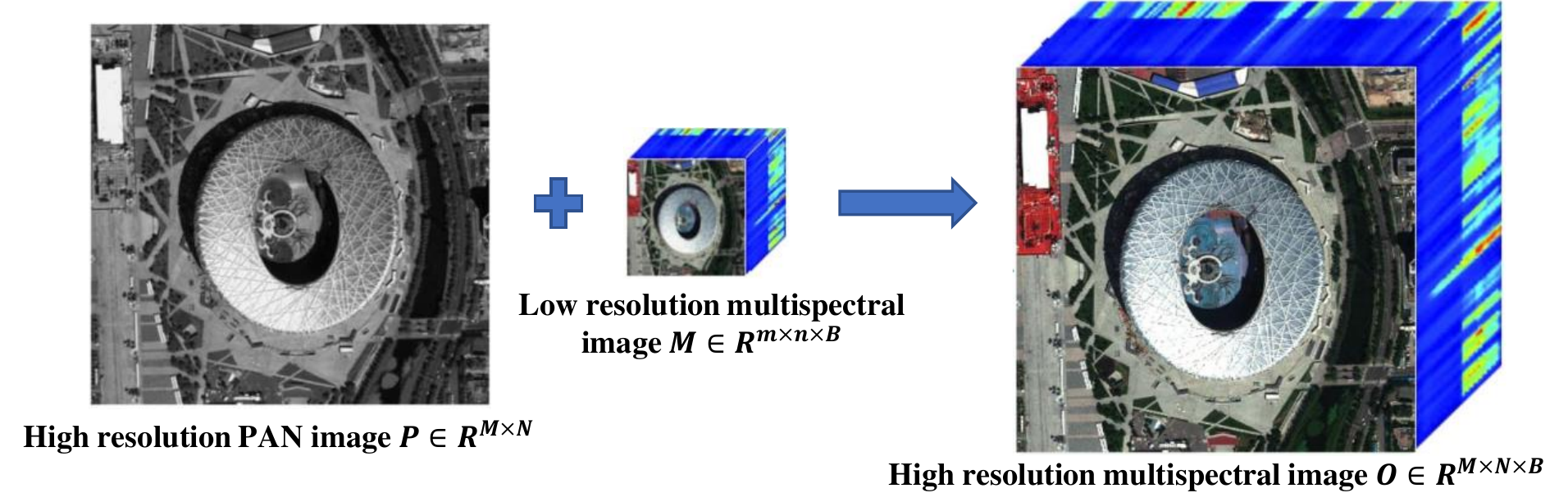}
	\caption{The schematic diagram of pansharpening.}\label{intro_sharpening}
\end{figure}

Pansharpening is an important task for remote sensing image processing. It aims at obtaining a HRMS image by fusing a PAN image with a LRMS image as shown in Figure~\ref{intro_sharpening}. With the satellite technology developing rapidly, the PAN and LRMS images can be easily acquired simultaneously by sensors equipped on the satellites, which requires handling the pansharpening task efficiently. Recently, a large number of approaches have been proposed, which can be roughly divided into four categories, i.e., component substitution (CS) approaches, multiresolution analysis (MRA) methods, variational optimization (VO) techniques, and deep learning (DL) approaches~\cite{vivone2014critical}.

The CS-based approaches rely on the concept of the projection of the LRMS image into a new domain, where the spatial information can be easily separated into a component, usually called the intensity component. Then, the PAN image can
be substituted with the intensity component. Representative methods include principal component analysis (PCA)~\cite{shah2008efficient}, intensity–hue-saturation~\cite{carper1990use}, and Gram–Schmidt spectral sharpening appoach~\cite{laben2000process}. The CS-based methods are easy to implement and can approximate the resolution of the HRMS image in the spatial domain but result in spectral distortion~\cite{vivone2014critical}. To alleviate this issue, some improved methods have been presented, such as the spatial-detail approach with the band-dependent property (BDSD)~\cite{garzelli2007optimal}, the robust version of BDSD approach~\cite{vivone2019robust}, and the adaptive CS approach with partial replacement~\cite{choi2010new}. 

The MRA-based methods depend on injecting the spatial details captured by decomposing the PAN image in a multiresolution way to the LRMS image. This type of method can preserve spectral information but bring in spatial distortion. Typical methods include the intensity modulation with smooth filter~\cite{liu2000smoothing}, the additive wavelet luminance proportional (AWLP)~\cite{otazu2005introduction}, the Laplacian pyramid-based methods~\cite{restaino2019pansharpening,vivone2020pansharpening,vivone2018full}. 

The VO-based techniques are adopted by formulating pansharpening as a variational optimization problem, which contains elegant mathematical derivation but often suffers from a high computational burden. Typical approaches include the Bayesian methods~\cite{wang2018high}, variational approaches~\cite{fang2013variational,fang2013variational2,deng2018variational,deng2019fusion,fu2019variational,tian2021variational}, and compressed sensing approaches~\cite{li2010new}. 

Recently, the DL-based approaches have been extensively studied, and this type of method directly learns the mapping from LRMS and PAN to HRMS by some stacked network architectures. For example, Huang et al.~\cite{huang2015new} firstly proposed a stacked modified sparse denoising autoencoder (S-MSDA) to train the relationship between high-resolution (HR) and low-resolution (LR) image patches. Masi et al.~\cite{masi2016pansharpening} designed a simple and effective three-layer convolutional neural network (CNN) for the pansharpening problem. Further, a deeper network based on the ResNet is built in~\cite{yang2017pannet,fu2020deep} by focusing on the spectral and spatial preservations. A multiple-scale and multiple-depth CNN is proposed in~\cite{yuan2018multiscale}. Later, He et al.~\cite{he2019pansharpening} proposed a network by combining the detail injection strategy with the CNN. Lately, Xie et al.~\cite{xie2020hpgan} designed a generative adversarial network (GAN) based network. Zhou et al.~\cite{zhou2021pgman} proposed an unsupervised generative multiadversarial network. Similarly, Deng et al.~\cite{deng2020detail} devised a detailed injection-based network by deriving from the CS and MRA frameworks. More recently, Yang et al.~\cite{yang2021dual} proposed a dual-stream CNN with residual information enhancement. Zhang et al.~\cite{zhang2021gtp} developed a residual learning network based on gradient transformation prior. However, the network architectures of these DL-based approaches are empirically designed by stacking some network modules, and thus the entire architectures can be seen as black-box mechanisms, and the role of different modules in these networks is difficult to analyze and understand. To alleviate this issue, some attempts by combining the model-based methodology with the deep learning method have been made. For example, Xie et al.~\cite{xie2020mhf} designed an interpretable deep network for multispectral and hyperspectral image fusion. Dong et al.~\cite{dong2021model} proposed a model-guided deep network for hyperspectral image super-resolution. Yin et al.~\cite{yin2021pscsc} and Cao et al.~\cite{cao2021pancsc} proposed two model-driven deep networks for pansharpening.

In this paper, we keep exploring in this direction and propose a novel deep network, dubbed as \textit{Proximal PanNet}, for the pansharpening problem by combining the model-based methodology with the deep learning method. Firstly, we propose a new observation model for pansharpening using the convolutional sparse coding (CSC) technique. Specifically, we model the PAN and MS image separately using the CSC with two types of features, i.e., common features and unique features. The two features are based on two observations. One is that the PAN and MS images capture the same scene, they thus share some common features, and the other one is that the two images are generated by different sensors equipped on the satellite, they thus contain their unique features. Therefore, to generate a better fused HRMS image, we need to capture the common and unique features from the two images, and then reconstruct the HRMS image using both features. Secondly, we reformulate the observation model into an optimization problem and add implicit regularizations modeling priors on both common and unique features. Thirdly, we design a proximal gradient algorithm to solve this optimization model, and unfold the iterative algorithm into a deep network, where each module corresponds to a specific operation of the iterative algorithm. The proximal operators are approximated by the convolutional neural network. Therefore, each module of the unfolded network can be clearly interpreted. Finally, all the learnable modules can be automatically learned end-to-end, and thus the hyperparameter selection can be avoided. Experimental results show the superiority of our network both quantitatively and qualitatively comparing with other state-of-the-art methods.


\section{Proposed Model}
In this section, a new pansharpening model is firstly proposed. Then, a proximal gradient algorithm is derived to optimize this model. Finally, the algorithm is further unfolded into a deep neural network.

\subsection{Proposed Model for Pansharpening}
The pansharpening model is proposed based on the following observations. Firstly, since both the LRMS image $\textbf{M}\in\mathbb{R}^{m\times n\times B}$ and the PAN image $\textbf{P}\in\mathbb{R}^{M\times N}$ are captured in the same scene, they should share some common features. Secondly, since the two images are acquired by different sensors in the satellite, each image thus has its unique feature. Additionally, we also assume that the information underlying both the LRMS image and the PAN image are beneficial to estimate the HRMS image $\textbf{O}\in\mathbb{R}^{M\times N\times B}$. Therefore, we firstly extract these effective features from both images, and then fuse these features to generate the HRMS image $\textbf{O}$. To extract the features from both images, we adopt the CSC technique~\cite{deng2020deep,li2021online} to model the PAN and LRMS images separately. 

Specifically, our new model is defined as follows:
\begin{eqnarray}
\mathbf{P} &=& \sum_{k=1}^{K}\mathbf{D}_{k}^{c}\otimes\mathbf{C}_{k} + \sum_{k=1}^{K}\mathbf{D}_{k}^{u}\otimes\mathbf{U}_{k},\\
\tilde{\mathbf{M}} &=& \sum_{k=1}^{K}\mathbf{H}_{k}^{c}\otimes\mathbf{C}_{k} + \sum_{k=1}^{K}\mathbf{H}_{k}^{v}\otimes\mathbf{V}_{k},
\end{eqnarray}
where $\tilde{\textbf{M}}\in \mathbb{R}^{M\times N\times B}$ is the upsampled version of the LRMS image $\textbf{M}$ using the polynomial kernel technique~\cite{aiazzi2002context}, $\otimes$ is the convolutional operation, $\mathbf{D}_{k}^{c}\in\mathbb{R}^{s\times s}$ and $\mathbf{H}_{k}^{c}\in\mathbb{R}^{s\times s\times B}$ denote the common filter for both images;  $\mathbf{D}_{k}^{u}\in\mathbb{R}^{s\times s}$ and $\mathbf{H}_{k}^{v}\in\mathbb{R}^{s\times s\times B}$ represent the separate filters of both images; $\mathbf{C}_{k}\in\mathbb{R}^{M\times N}$ denotes the shared common features, and $\mathbf{U}_{k}\in\mathbb{R}^{M\times N}$ and $\mathbf{V}_{k}\in\mathbb{R}^{M\times N}$ are unique features of HRMS image and LRHS image, respectively. Then, the HRMS image $\mathbf{O}\in\mathbb{R}^{M\times N\times B}$ can be generated by fusing these features as follows:
\begin{eqnarray}
\mathbf{O} \!=\! \sum_{k=1}^{K}\mathbf{G}_{k}^{c}\!\otimes\!\mathbf{C}_{k} \!+\! \sum_{k=1}^{K}\mathbf{G}_{k}^{u}\!\otimes\!\mathbf{U}_{k}\!+\! \sum_{k=1}^{K}\mathbf{G}_{k}^{v}\!\otimes\!\mathbf{V}_{k},
\end{eqnarray}
where $\mathbf{G}_{k}^{c}\in\mathbb{R}^{s\times s\times B}$ represents the common filter, $\mathbf{G}_{k}^{u}\in\mathbb{R}^{s\times s\times B}$ and $\mathbf{G}_{k}^{v}\in\mathbb{R}^{s\times s\times B}$ denote the unique filters concerning the features of both images. To make the notations clear, we denote $\mathbf{C}=\{\mathbf{C}_{k}\}_{k=1}^{K}\in\mathbb{R}^{M\times N\times K}$,  $\mathbf{U}=\{\mathbf{U}_{k}\}_{k=1}^{K}\in\mathbb{R}^{M\times N\times K}$, and $\mathbf{V}=\{\mathbf{V}_{k}\}_{k=1}^{K}\in\mathbb{R}^{M\times N\times K}$.

In the following, we assume that all the filters are known. Therefore, to obtain the fused HRMS image $\textbf{O}$, we need to calculate $\mathbf{C}$, $\mathbf{U}$, and $\mathbf{V}$ as follows:
\begin{equation}\label{ori_model}
\begin{split}
\min_{\mathbf{C}, \mathbf{U}, \mathbf{V}}\frac{1}{2}||\mathbf{P} - \sum_{k=1}^{K}\mathbf{D}_{k}^{c}\otimes\mathbf{C}_{k} - \sum_{k=1}^{K}\mathbf{D}_{k}^{u}\otimes\mathbf{U}_{k}||_{F}^{2}\\
+~\frac{1}{2}||\tilde{\mathbf{M}} -\sum_{k=1}^{K}\mathbf{H}_{k}^{c}\otimes\mathbf{C}_{k} - \sum_{k=1}^{K}\mathbf{H}_{k}^{v}\otimes\mathbf{V}_{k}||_{F}^{2} \\
+\lambda_{1}f_{1}(\mathbf{U})+\lambda_{2}f_{2}(\mathbf{V})+\lambda_{3}f_{3}(\mathbf{C}),
\end{split}
\end{equation}
where $\lambda_1, \lambda_2$, and $\lambda_3$ are the trade-off parameters. $f_{1}(\cdot), f_{2}(\cdot)$, and $f_{3}(\cdot)$ are the regularization terms modeling the priors on $\mathbf{U}$, $\mathbf{V}$, and $\mathbf{C}$, respectively.

\subsection{Model Optimization}
To solve the problem, we adopt the alternate optimization algorithm to alternately update each variable with other variables fixed, which leads to the three sub-problems:
\begin{eqnarray}
\begin{split}
\mathbf{U}^{(t)}\!=\!\arg\!\min_{\mathbf{U}}&||\mathbf{P}\!-\! \sum_{k=1}^{K}\!\mathbf{D}_{k}^{c}\!\otimes\!\mathbf{C}_{k}^{(t-1)} \!-\! \sum_{k=1}^{K}\!\mathbf{D}_{k}^{u}\!\otimes\!\mathbf{U}_{k}||_{F}^{2}\!\\&+\!\lambda_{1}f_{1}(\mathbf{U}),
\end{split}
\end{eqnarray}
\vspace{-0.3cm}
\begin{eqnarray}
\begin{split}
\mathbf{V}^{(t)}\!=\!\arg\!\min_{\mathbf{V}}&||\tilde{\mathbf{M}} \!\!-\!\!\sum_{k=1}^{K}\mathbf{H}_{k}^{c}\!\!\otimes\!\!\mathbf{C}_{k}^{(t-1)} \!\!-\!\! \sum_{k=1}^{K}\mathbf{H}_{k}^{v}\!\!\otimes\!\!\mathbf{V}_{k}||_{F}^{2}\!\\&+\!\lambda_{2}f_{2}(\mathbf{V}),
\end{split}
\end{eqnarray}
\vspace{-0.3cm}
\begin{eqnarray}
\begin{split}
&\mathbf{C}^{(t)}\!=\!\arg\!\min_{\mathbf{C}}||\mathbf{P} \!\!-\!\! \sum_{k=1}^{K}\mathbf{D}_{k}^{c}\!\otimes\!\mathbf{C}_{k} \!-\! \sum_{k=1}^{K}\mathbf{D}_{k}^{u}\!\otimes\!\mathbf{U}_{k}^{(t)}||_{F}^{2}\\
&\!+\!||\tilde{\mathbf{M}} \!-\!\sum_{k=1}^{K}\mathbf{H}_{k}^{c}\!\otimes\!\mathbf{C}_{k} \!\!-\!\! \!\sum_{k=1}^{K}\mathbf{H}_{k}^{v}\!\otimes\!\mathbf{V}_{k}^{(t)}||_{F}^{2}\!+\!\lambda_{3}f_{3}(\mathbf{C}).
\end{split}
\end{eqnarray}
Next, we introduce the optimization for each sub-problem.

\subsubsection{$\mathbf{U}-$subproblem}
Instead of directly optimizing the objective (5), we first adopt the quadratic function to approximate the objective function, which is expressed as:
\begin{eqnarray}\label{u-subproblem}
\begin{split}
\mathbf{U}^{(t)}\!=\!\arg\min_{\mathbf{U}}g\left(\mathbf{U}^{(t-1)}\right)\!+\!\frac{1}{2\eta_1}||\mathbf{U}-\mathbf{U}^{(t-1)}||_{F}^{2}\\
\!+\!\left<\nabla g(\mathbf{U}^{(t-1)}), \mathbf{U}\!-\!\mathbf{U}^{(t-1)}\right> \!+\! \lambda_{1}f_{1}(\mathbf{U}),
\end{split}
\end{eqnarray}
where 
\vspace{-0.3cm}
\begin{equation}
g\left(\mathbf{U}^{(t\!-\!1)}\right)\!\!\!=\!\!||\mathbf{P}\!\!-\!\!\! \sum_{k=1}^{K}\!\mathbf{D}_{k}^{c}\!\otimes\!\mathbf{C}_{k}^{(t\!-\!1)} \!\!-\!\!\! \sum_{k=1}^{K}\!\mathbf{D}_{k}^{u}\!\otimes\!\mathbf{U}_{k}^{(t\!-\!1)}\!\!||_{F}^{2},
\end{equation}
\begin{equation}
\nabla g(\mathbf{U}^{(t\!-\!1)})\!\!=\!\!\mathbf{D}^{u}\!\otimes^{\dagger}\!\left( \sum_{k=1}^{K}\!\mathbf{D}_{k}^{c}\!\otimes\!\mathbf{C}_{k}^{(t\!-\!1)} \!\!+\!\! \sum_{k=1}^{K}\!\mathbf{D}_{k}^{u}\!\otimes\!\mathbf{U}_{k}^{(t\!-\!1)}\!\!-\!\!\mathbf{P}\right),
\end{equation}
$\nabla(\cdot)$ is gradient operator, $\eta_1$ is the step size, $\mathbf{D}^{u}\in\mathbb{R}^{s\times s\times K\times 1}$ is a 4$-$D tensor stacked by all $\mathbf{D}_{k}^{u}$, and $\otimes^{\dagger}$ denotes the transposed convolution. By a simple mathematical derivation, Eq. (\ref{u-subproblem}) can be equivalent to
\begin{eqnarray}\label{u-sub2}
\begin{split}
\mathbf{U}^{(t)}=\arg\min_{\mathbf{U}}&\frac{1}{2}||\mathbf{U}-(\mathbf{U}^{(t-1)}-\eta_{1}\nabla g(\mathbf{U}^{(t-1)}))||_{F}^{2}\\
&+\eta_{1}\lambda_{1}f_{1}(\mathbf{U}).
\end{split}
\end{eqnarray}
Eq.(\ref{u-sub2}) can then be optimized by the general proximal operator~\cite{beck2009fast}, which is defined as
\begin{eqnarray}\label{eq.solveU}
\mathbf{U}^{(t)}=prox_{\eta_{1}\lambda_{1}}\left(\mathbf{U}^{(t-1)}-\eta_{1}\nabla g(\mathbf{U}^{(t-1)})\right),
\end{eqnarray}
where $prox_{\eta_{1}\lambda_{1}}(\cdot)$ is the proximal operator related to $f_{1}(\cdot)$. To alleviate the limitation of human designed regularizers, we adopt the deep CNNs to automatically learn the prior from the data, which will be discussed in detail in the next subsection.

\subsubsection{$\mathbf{V}-$subproblem}
For the updating of $\mathbf{V}$, we adopt a similar strategy. Firstly, the quadratic approximation of (6) with respect to $\mathbf{V}$ is defined as:
\begin{eqnarray}\label{v-subproblem}
\begin{split}
\mathbf{V}^{(t)}=\arg\min_{\mathbf{V}}h\left(\mathbf{V}^{(t-1)}\right)+\frac{1}{2\eta_2}||\mathbf{V}-\mathbf{V}^{(t-1)}||_{F}^{2}\\
+\left<\nabla h(\mathbf{V}^{(t-1)}), \mathbf{V}-\mathbf{V}^{(t-1)}\right>+\lambda_{2}f_{2}(\mathbf{V}),
\end{split}
\end{eqnarray}
where 
\vspace{-0.3cm}
\begin{equation}
h\left(\mathbf{V}^{(t-1)}\right)\!\!=\!\!||\tilde{\mathbf{M}} \!\!-\!\! \sum_{k=1}^{K}\!\mathbf{H}_{k}^{c}\!\otimes\!\mathbf{C}_{k}^{(t-1)} \!\!-\!\! \sum_{k=1}^{K}\!\mathbf{H}_{k}^{v}\!\otimes\!\mathbf{V}_{k}^{(t-1)}||_{F}^{2},
\end{equation}
\begin{equation}
\nabla h(\mathbf{V}^{(t-1)})\!\!=\!\!\mathbf{H}^{v}\!\!\otimes^{\dagger}\!\!\left( \sum_{k=1}^{K}\!\mathbf{H}_{k}^{c}\!\!\otimes\!\!\mathbf{C}_{k}^{(t\!-\!1)} \!\!+\!\! \sum_{k=1}^{K}\!\mathbf{H}_{k}^{v}\!\otimes\!\mathbf{V}_{k}^{(t\!-\!1)}\!\!-\!\!\tilde{\mathbf{M}}\right),
\end{equation}
$\eta_2$ is the step size, and $\mathbf{H}^{v}\in\mathbb{R}^{s\times s\times K\times B}$ is a 4$-$D tensor stacked by all $\mathbf{H}_{k}^{v}$. Further, Eq. (\ref{v-subproblem}) can be equivlently formulated as
\begin{eqnarray}\label{v-sub2}
\begin{split}
\mathbf{V}^{(t)}=\arg\min_{\mathbf{V}}&\frac{1}{2}||\mathbf{V}-(\mathbf{V}^{(t-1)}-\eta_{2}\nabla h(\mathbf{V}^{(t-1)}))||_{F}^{2}\\
&+\eta_{2}\lambda_{2}f_{2}(\mathbf{V}),
\end{split}
\end{eqnarray}
Then, Eq.(\ref{v-sub2}) can also be solved by
\begin{eqnarray}\label{eq.solveV}
\mathbf{V}^{(t)}=prox_{\eta_{2}\lambda_{2}}\left(\mathbf{V}^{(t-1)}-\eta_{2}\nabla h(\mathbf{V}^{(t-1)})\right),
\end{eqnarray}
where $prox_{\eta_{2}\lambda_{2}}(\cdot)$ is the proximal operator related to $f_{2}(\cdot)$.

\subsubsection{$\mathbf{C}-$subproblem} Eq. (7) can be re-written as:
\begin{eqnarray}\label{C-subproblem}
\begin{split}
\mathbf{C}^{(t)}&=\arg\min_{\mathbf{C}}||\hat{\mathbf{P}} - \sum_{k=1}^{K}\mathbf{D}_{k}^{c}\otimes\mathbf{C}_{k}||_{F}^{2}\\
&+||\hat{\mathbf{M}} -\sum_{k=1}^{K}\mathbf{H}_{k}^{c}\otimes\mathbf{C}_{k}||_{F}^{2} +\lambda_{3}f_{3}(\mathbf{C}),
\end{split}
\end{eqnarray}
where $\hat{\mathbf{P}}=\mathbf{P} - \sum_{k=1}^{K}\mathbf{D}_{k}^{u}\otimes\mathbf{U}_{k}^{(t)}$ and $\hat{\mathbf{M}}=\tilde{\mathbf{M}} - \sum_{k=1}^{K}\mathbf{H}_{k}^{v}\otimes\mathbf{V}_{k}^{(t)}$. Further, Eq. (\ref{C-subproblem}) can be expressed as:
\begin{eqnarray}\label{C-sub2}
\mathbf{C}^{(t)}\!=\!\arg\min_{\mathbf{C}}||\mathbf{N} \!\!-\!\! \sum_{k=1}^{K}\mathbf{L}_{k}^{c}\otimes\mathbf{C}_{k}||_{F}^{2} \!+\!\lambda_{3}f_{3}(\mathbf{C}),
\end{eqnarray}
where $\mathbf{N}\in\mathbb{R}^{M\times N\times (B+1)}$ is the concatenation of $\hat{\mathbf{P}}$ and $\hat{\mathbf{M}}$, and $\mathbf{L}_{k}^{c}\in\mathbb{R}^{s\times s\times (B+1)}$ is the concatenation of $\mathbf{D}_{k}^{c}$ and $\mathbf{H}_{k}^{c}$.
Eq. (\ref{C-sub2}) can then be solved by the same way as optimizing $\mathbf{U}$ and $\mathbf{V}$. More specifically, we first derive the quadratic approximation of Eq. (\ref{C-sub2}) with respect to $\mathbf{C}$ as follows:
\begin{eqnarray}\label{c-sub3}
\begin{split}
\mathbf{C}^{(t)}=\arg\min_{\mathbf{C}}l\left(\mathbf{C}^{(t-1)}\right)+\frac{1}{2\eta_3}||\mathbf{C}-\mathbf{C}^{(t-1)}||_{F}^{2}\\
+\left<\nabla l(\mathbf{C}^{(t-1)}), \mathbf{C}-\mathbf{C}^{(t-1)}\right>+\lambda_{3}f_{3}(\mathbf{C}),
\end{split}
\end{eqnarray}
where
\vspace{-0.3cm} 
\begin{eqnarray}
l\left(\mathbf{C}^{(t-1)}\right)\!\!&=&\!\!||\mathbf{N} \!\!-\!\! \sum_{k=1}^{K}\!\mathbf{L}_{k}^{c}\!\otimes\!\mathbf{C}_{k}^{(t-1)} ||_{F}^{2},\\
\nabla l(\mathbf{C}^{(t-1)})\!\!&=&\!\!\mathbf{L}^{c}\!\otimes^{\dagger}\!\left( \sum_{k=1}^{K}\!\mathbf{L}_{k}^{c}\!\otimes\!\mathbf{C}_{k}^{(t-1)} \!-\!\mathbf{N}\right),
\end{eqnarray}
$\eta_3$ is the step size, and $\mathbf{L}^{c}\in\mathbb{R}^{s\times s\times K\times (B+1)}$ is a 4$-$D tensor stacked by all $\mathbf{L}_{k}^{c}$. Further, Eq. (\ref{c-sub3}) can be rewritten as 
\begin{eqnarray}\label{c-sub4}
\begin{split}
\mathbf{C}^{(t)}=\arg\min_{\mathbf{C}}&\frac{1}{2}||\mathbf{C}-(\mathbf{C}^{(t-1)}-\eta_{3}\nabla l(\mathbf{C}^{(t-1)}))||_{F}^{2}\\
&+\eta_{3}\lambda_{3}f_{3}(\mathbf{C}).
\end{split}
\end{eqnarray}
Finally, Eq.(\ref{c-sub4}) can be solved as
\begin{eqnarray}\label{eq.solveC}
\mathbf{C}^{(t)}=prox_{\eta_{3}\lambda_{3}}\left(\mathbf{C}^{(t-1)}-\eta_{3}\nabla l(\mathbf{C}^{(t-1)})\right),
\end{eqnarray}
where $prox_{\eta_{3}\lambda_{3}}(\cdot)$ is the proximal operator related to $f_{3}(\cdot)$.

\begin{figure}
	\centering
	\includegraphics[width=1\linewidth]{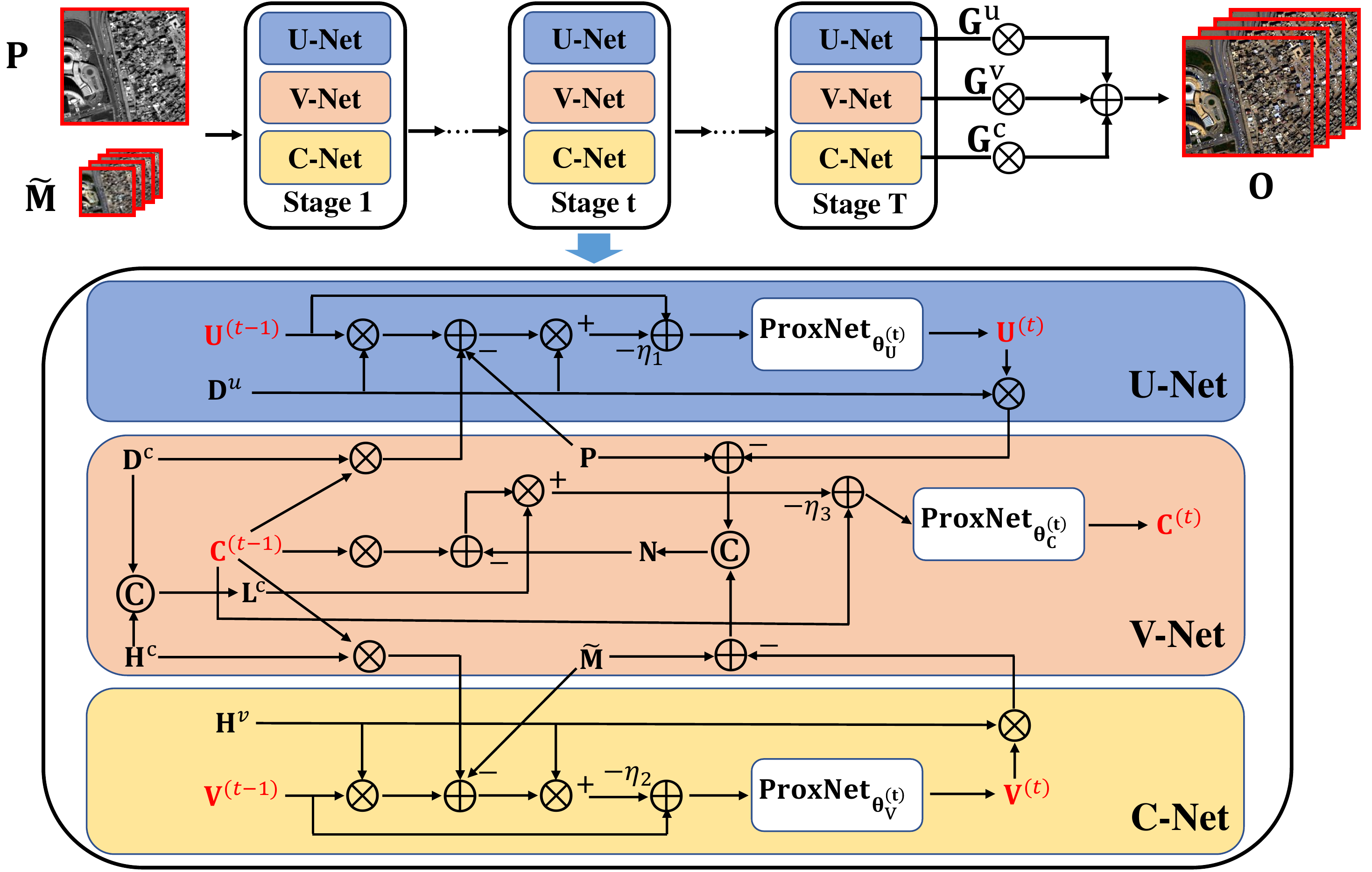}
	\caption{The overall architecture of our proposed unfolding network for pansharpening. This network is composed of $T$ stages shown at the top, and each stage contains three sub-networks (i.e., U-Net, V-Net, and C-net). The detailed network architectures of the three sub-networks in the stage $t$ are shown at the bottom. $\otimes$ and $\oplus$ represent the convolutional operation and the element-wise addition, respectively.}\label{network}
\end{figure}

\subsection{Proximal PanNet}
Based on the iterative algorithm, we build a deep network shown in the top row of Fig.~\ref{network}. This network contains $T$ stages corresponding to $T$ iterations of the iterative algorithm for solving Eq. (\ref{ori_model}). Each stage takes the outputs of the previous stage $\mathbf{U}^{(t-1)}$, $\mathbf{V}^{(t-1)}$ and $\mathbf{C}^{(t-1)}$ as the inputs and produces the updates $\mathbf{U}^{(t)}$, $\mathbf{V}^{(t)}$ and $\mathbf{C}^{(t)}$.

As shown in Eq. (\ref{eq.solveU}),  (\ref{eq.solveV}) and (\ref{eq.solveC}), the three proximal operators ${\rm{pro}}{{\rm{x}}_{{\eta _1}{\lambda _1}}}(\cdot)$, ${\rm{pro}}{{\rm{x}}_{{\eta _2}{\lambda _2}}}(\cdot)$,  and ${\rm{pro}}{{\rm{x}}_{{\eta _3}{\lambda _3}}}(\cdot)$ play a key role in the algorithm. To update each variable, classical methods usually design hand-crafted priors which make the proximal operators be closed form. However, due to the limited representation abilities, human-designed priors cannot always achieve the expected effect. Therefore, we utilize convolutional neural networks to explore prior for $\mathbf{U}$, $\mathbf{V}$ and $\mathbf{C}$, namely learning their corresponding proximal operators ${\rm{pro}}{{\rm{x}}_{{\eta _1}{\lambda _1}}}(\cdot)$, ${\rm{pro}}{{\rm{x}}_{{\eta _2}{\lambda _2}}}(\cdot)$,  and ${\rm{pro}}{{\rm{x}}_{{\eta _3}{\lambda _3}}}(\cdot)$ for updating $\mathbf{U}^{(t)}$, $\mathbf{V}^{(t)}$, and $\mathbf{C}^{(t)}$ in each stage $t$.

\subsubsection{Network design}
Our network is designed as several stages, and each stage updates $\mathbf{U}$, $\mathbf{V}$ and $\mathbf{C}$ in a cascaded manner. At stage $t$, $\mathbf{U}^{(t)}$ is updated using $\mathbf{P}$, $\mathbf{U}^{(t-1)}$, and $\mathbf{C}^{(t-1)}$ by the following sub-steps:
\vspace{-0.2cm}
\begin{align}
\begin{split}
\mathbf{U-Net}:\left\{
\begin{array}{ll}\label{eq.netU}
&\!\!\!\!\!\mathbf{P}_{c}^{(t-1)}=\sum_{k=1}^{K}\mathbf{D}_{k}^{c}\otimes\mathbf{C}_{k}^{(t-1)},  \\
&\!\!\!\!\!\mathbf{P}_{u}^{(t-1)}=\sum_{k=1}^{K}\mathbf{D}_{k}^{u}\otimes\mathbf{U}_{k}^{(t-1)},   \\
&\!\!\!\!\!\mathbf{\varepsilon}_{P}^{(t-1)} = \mathbf{P}_{c}^{(t-1)} + \mathbf{P}_{u}^{(t-1)} - \mathbf{P},  \\
&\!\!\!\!\!\nabla g^{(t-1)}=\mathbf{D}^{u}\otimes^{\dagger}\mathbf{\varepsilon}_{P}^{(t-1)},  \\
&\!\!\!\!\!\mathbf{U}^{(t-0.5)}=\mathbf{U}^{(t-1)} - \eta_{1}\nabla g^{(t-1)},  \\
&\!\!\!\!\!\mathbf{U}^{(t)} = {\rm{proxNe}}{{\rm{t}}_{\Theta _{\bf{U}}^{(t)}}}\left( \mathbf{U}^{(t-0.5)}\right).
\end{array}
\right.
\end{split}
\end{align}
Each step is unfolded into one module of the U-Net shown in Figure~\ref{network}. Then, $\mathbf{V}^{(t)}$ is updated by $\tilde{\mathbf{M}}$, $\mathbf{V}^{(t-1)}$, and $\mathbf{C}^{(t-1)}$:
\begin{align}
\begin{split}
\mathbf{V-Net}: \left\{
\begin{array}{ll}\label{eq.netV}
&\!\!\!\!\!\mathbf{M}_{c}^{(t-1)}=\sum_{k=1}^{K}\mathbf{H}_{k}^{c}\otimes\mathbf{C}_{k}^{(t-1)},  \\
&\!\!\!\!\!\mathbf{M}_{v}^{(t-1)}=\sum_{k=1}^{K}\mathbf{H}_{k}^{v}\otimes\mathbf{V}_{k}^{(t-1)},   \\
&\!\!\!\!\!\mathbf{\varepsilon}_{M}^{(t-1)} = \mathbf{M}_{c}^{(t-1)} + \mathbf{\textsc{M}}_{v}^{(t-1)} - \tilde{\mathbf{M}},  \\
&\!\!\!\!\!\nabla h^{(t-1)}=\mathbf{H}^{v}\otimes^{\dagger}\mathbf{\varepsilon}_{M}^{(t-1)},  \\
&\!\!\!\!\!\mathbf{V}^{(t-0.5)} =  \mathbf{V}^{(t-1)} - \eta_{2}\nabla h^{(t-1)}, \\
&\!\!\!\!\!\mathbf{V}^{(t)} = {\rm{proxNe}}{{\rm{t}}_{\Theta _{\bf{V}}^{(t)}}}\left(\mathbf{V}^{(t-0.5)}\right).
\end{array}
\right.
\end{split}
\end{align}
Similarly, each step is unfolded into one module of the V-Net as shown in Figure~\ref{network}. Finally, $\mathbf{C}^{(t)}$ is updated using $\mathbf{P}$, $\tilde{\mathbf{M}}$, $\mathbf{C}^{(t-1)}$, $\mathbf{U}^{(t)}$ and $\mathbf{V}^{(t)}$:
\begin{align}
\begin{split}
\mathbf{C-Net}:\left\{
\begin{array}{ll}
&\!\!\!\!\!\mathbf{F}_{c}^{(t-1)}=\sum_{k=1}^{K}\mathbf{L}_{k}^{c}\otimes\mathbf{C}_{k}^{(t-1)},   \\
&\!\!\!\!\!\mathbf{\varepsilon}_{C}^{(t-1)} = \mathbf{F}_{c}^{(t-1)} - \mathbf{N},  \\
&\!\!\!\!\!\nabla l^{(t-1)}=\mathbf{L}^{c}\otimes^{\dagger}\mathbf{\varepsilon}_{C}^{(t-1)},  \\
&\!\!\!\!\!\nabla l^{(t-0.5)}=\mathbf{C}^{(t-1)} - \eta_{3}\nabla l^{(t-1)},  \\
&\!\!\!\!\!\mathbf{C}^{(t)} = {\rm{proxNe}}{{\rm{t}}_{\Theta _{\bf{C}}^{(t)}}}\left(l^{(t-0.5)}\right).
\end{array}
\right.
\end{split}
\end{align}
Similarly, each step is unfolded into one module of the C-Net as shown in Figure~\ref{network}. Additionally,  ${\rm{proxNe}}{{\rm{t}}_{\Theta _{\bf{U}}^{(t)}}}(\cdot)$, ${\rm{proxNe}}{{\rm{t}}_{\Theta _{\bf{V}}^{(t)}}}(\cdot)$ and ${\rm{proxNe}}{{\rm{t}}_{\Theta _{\bf{C}}^{(t)}}}(\cdot)$ are all ResNet~\cite{He2016Deep} with 3 ResBlocks, which are used to approximate the three proximal operators, and ${\Theta _{\bf{U}}^{(t)}}$, ${\Theta _{\bf{V}}^{(t)}}$, and ${\Theta _{\bf{C}}^{(t)}}$ are the corresponding network parameters at the $t^{th}$ stage. Besides, ${\Theta _{\bf{U}}^{(t)}}$, ${\Theta _{\bf{V}}^{(t)}}$, and ${\Theta _{\bf{C}}^{(t)}}$ are the parameters of the three ResNets at the $t^{th}$ stage. Note that all the network parameters can be end-to-end learned from training data, including $\{\Theta_{\bf{U}}^{(t)}, \Theta_{\bf{V}}^{(t)}, \Theta_{\bf{C}}^{(t)}\}_{t=1}^{T}$, $\mathbf{D}^{c}=\{\mathbf{D}^{c}_{k}\}_{k=1}^{K}$, $\mathbf{D}^{u}=\{\mathbf{D}^{u}_{k}\}_{k=1}^{K}$, $\mathbf{H}^{c}=\{\mathbf{H}^{c}_{k}\}_{k=1}^{K}$, $\mathbf{H}^{v}=\{\mathbf{H}^{v}_{k}\}_{k=1}^{K}$, $\mathbf{G}^{c}=\{\mathbf{G}^{c}_{k}\}_{k=1}^{K}$, $\mathbf{G}^{u}=\{\mathbf{G}^{u}_{k}\}_{k=1}^{K}$, $\mathbf{G}^{v}=\{\mathbf{G}^{v}_{k}\}_{k=1}^{K}$, $\eta_1$, $\eta_2$, and $\eta_3$. Once training finished, the testing can be shortened due to its feed-forward architecture. Note that our network shown in Figure \ref{network} is interpretable. For example,To update $\mathbf{U}^{(t)}$ by Eq. (\ref{eq.netU}), computing $\mathbf{U}^{(t-0.5)}$ means the gradient descent, and the ${\rm{proxNe}}{{\rm{t}}_{\Theta _{\bf{U}}^{(t)}}}(\cdot)$ performs nonlinear operation on ${\bf{U}}^{(t-0.5)}$.

\subsubsection{Network training} 
The loss function of our training process is defined as 
\vspace{-0.3cm}
\begin{eqnarray}
\mathcal{L}=\sum_{j=1}^{J}||\mathbf{O}_{p}^{j}-\mathbf{O}_{g}^{j}||_{2}^{2},
\end{eqnarray}
where $\mathbf{O}_{p}^{j}$ is the predicted HRMS image of the $j^{th}$ training sample, $\mathbf{O}_{g}^{j}$ is the true HRMS image of the $j^{th}$ training sample, and $J$ is the number of training pairs. In our network, all the convolutional kernel sizes are set as 8$\times$8. The channel number of each convolution is set as 16. The stage number of our network is set as 2. We implement our network using TensorFlow on a GTX 1080Ti GPU with 12 GB memory. Additionally, we adopt the Adam algorithm, and a decayed technique to set the learning rate, i.e., decayed by 0.9 every 50 epochs with a fixed initial learning rate 0.0001. Also, the epoch number is 100, and the mini-batch size is 64.

\section{Experiments}
We conduct several experiments using data from the Worldview3 satellite. Specifically, we extract 12.5$K$ image pairs from the image of Worldview3, and each image pair contains one PAN image (64$\times$64), one MS image (16$\times$16$\times$8) and one ground truth HRMS image (64$\times$64$\times$8). We split these image pairs into 90\%/10\% for training/testing. The ground truth HRMS image is obtained by using the Wald’s protocol~\cite{wald1997fusion} due to its unavailability. We compare with twelve state-of-the-art methods: EXP~\cite{aiazzi2002context}, GS~\cite{laben2000process}, BDSD~\cite{garzelli2007optimal}, BDSD-PC~\cite{vivone2019robust}, PRACS~\cite{choi2010new}, AWLP~\cite{otazu2005introduction}, GLP~\cite{restaino2019pansharpening}, GLP-HPM~\cite{vivone2013contrast}, MO~\cite{restaino2016fusion}, SRD~\cite{vicinanza2014pansharpening}, PanNet~\cite{yang2017pannet}, and Fusion-Net~\cite{deng2020detail}. 

The performance assessment contains two cases. In the first case where the ground-truth HRMS is known, we adopt four popular reference-based indexes~\cite{vivone2014critical}: the Q4 (for 4 bands) and Q8 (for 8 bands), the spectral angle mapper (SAM), the relative dimensionless global error in synthesis (ERGAS), and the spatial correlation coefficient (SCC). In the second case where the ground-truth HRMS is not available, we use $D_{\lambda}$, $D_{s}$, and the quality without reference (QNR) criteria~\cite{vivone2014critical}. 

\subsection{Evaluation at Reduced Resolution}
We use 90\% of the image pairs to train our network and the remaining ones (1258 images) for testing. Table~\ref{result_rr} shows the average quantitative results and standard deviation of each method. From Table~\ref{result_rr}, it can be easily observed that the DL-based methods (PanNet, Fusion-Net, and our network) perform better than the CS and MRA approaches in terms of all the indexes, which may attribute to the strong non-linear approximation ability of the deep neural network. Additionally, compared with the PanNet and Fusion-Net, our proposed network obtains better performance, which further verifies that the CSC framework has a certain advantage over the CS and MRA frameworks since the PanNet and Fusion-Net are built based on the CS and MRA frameworks while our network is built based on the proposed CSC framework. Also, our proposed network is very robust since it has almost the smallest standard deviation.  

\begin{table}
	\setlength{\abovecaptionskip}{0cm}
	\setlength{\belowcaptionskip}{-0.2cm}
	\caption{\label{result_rr} Average quantitative results on the test dataset (Worldview3) in the reduced resolution case (1258 images).}
	\begin{center}
		{\small
			\scalebox{0.52}[0.52]{			
				\begin{tabular}{l|c c c c }
					\toprule
					\textbf{Methods} &\textbf{Q8}&\textbf{SAM}&\textbf{ERGAS}&\textbf{SCC}\\
					\midrule
					\textbf{EXP}~\cite{aiazzi2002context} & 0.6176$\pm$0.1614 & 5.7123$\pm$2.1141 & 6.7327$\pm$3.0030 & 0.8137$\pm$0.0694\\
					\textbf{GS}~\cite{laben2000process} & 0.8100$\pm$0.1301 & 6.7731$\pm$2.8316 & 5.0358$\pm$2.3855 & 0.9464$\pm$0.0329\\
					\textbf{BDSD}~\cite{garzelli2007optimal} & 0.8482$\pm$0.1274 & 5.3313$\pm$2.0875 & 4.1042$\pm$1.9688 & 0.9511$\pm$0.0328 \\
					\textbf{BDSD-PC}~\cite{vivone2019robust} & 0.7555$\pm$0.1495 & 5.6165$\pm$2.1201 & 5.1517$\pm$2.2935 & 0.9247$\pm$0.0394\\
					\textbf{PRACS}~\cite{choi2010new} & 0.8061$\pm$0.1400 & 5.4778$\pm$2.0946 & 4.5667$\pm$1.9784 & 0.9281$\pm$0.0405\\
					\textbf{AWLP}~\cite{otazu2005introduction} & 0.8479$\pm$0.1193 & 5.3395$\pm$1.9717 & 4.1996$\pm$1.9093 & 0.9476$\pm$0.0283\\
					\textbf{GLP}~\cite{restaino2019pansharpening} & 0.8533$\pm$0.1206 & 5.1268$\pm$1.9968 & 4.0173$\pm$1.8510 & 0.9506$\pm$0.0299\\
					\textbf{GLP-HPM}~\cite{vivone2013contrast} & 0.8472$\pm$0.1246 & 5.2111$\pm$2.0691 & 4.0717$\pm$1.8874 & 0.9458$\pm$0.0339\\
					\textbf{MO}~\cite{restaino2016fusion} & 0.8268$\pm$0.1327 & 5.2453$\pm$2.0049 & 4.4279$\pm$1.9556 & 0.9396$\pm$0.0317 \\
					\textbf{SRD}~\cite{vicinanza2014pansharpening} & 0.8324$\pm$0.1295 & 5.1725$\pm$2.0210 & 4.3116$\pm$1.9473 & 0.9418$\pm$0.0308 \\
					\textbf{PanNet}~\cite{yang2017pannet}& 0.9074$\pm$0.1258 & 3.9218$\pm$1.4958 & 2.6352$\pm$1.1725 & 0.9788$\pm$0.0235 \\ 
					\textbf{Fusion-Net}~\cite{deng2020detail} & \underline{0.9125$\pm$0.1220}  & \underline{3.8035$\pm$1.4406}  & \underline{2.5771$\pm$1.1607}  & \underline{0.9781$\pm$0.0202} \\
					\textbf{Our} & \textbf{0.9258$\pm$0.1203}  & \textbf{3.6421$\pm$1.4377}  & \textbf{2.4263$\pm$1.2355}  & \textbf{0.9850$\pm$0.0226} \\
					\midrule
					\textbf{Ideal value} & 1  & 0  & 0  & 1 \\
					\bottomrule		
				\end{tabular}	
			}
		}
	\end{center}	
\end{table} 

\begin{table}
	\setlength{\abovecaptionskip}{0cm}
	\setlength{\belowcaptionskip}{-0.2cm}
	\caption{\label{result_rr_1} Average quantitative results on the Tripoli dataset (Worldview3) in the reduced resolution case.}
	\begin{center}
		{\small
			\scalebox{0.7}[0.7]{			
				\begin{tabular}{l|c c c c }
					\toprule
					\textbf{Methods} &\textbf{Q8}&\textbf{SAM}&\textbf{ERGAS}&\textbf{SCC}\\
					\midrule
					\textbf{EXP}~\cite{aiazzi2002context} & 0.7032 & 6.7081 & 8.3318 & 0.7408\\
					\textbf{GS}~\cite{laben2000process} & 0.8158 & 7.2060 & 6.8264 & 0.8734\\
					\textbf{BDSD}~\cite{garzelli2007optimal} & 0.8334 & 6.5331 & 6.5594 & 0.8724\\
					\textbf{BDSD-PC}~\cite{vivone2019robust} & 0.7706 & 7.1170 & 7.1918 & 0.8605\\
					\textbf{PRACS}~\cite{choi2010new} & 0.8110 & 6.5833 & 6.8484 & 0.8584\\
					\textbf{AWLP}~\cite{otazu2005introduction} & 0.8434 & 6.3811 & 6.3808 & 0.8749\\
					\textbf{GLP}~\cite{restaino2019pansharpening} & 0.8464 & 6.2880 & 6.3224 & 0.8725\\
					\textbf{GLP-HPM}~\cite{vivone2013contrast} & 0.8406 & 6.3400 & 6.3929 & 0.8712\\
					\textbf{MO}~\cite{restaino2016fusion} & 0.8387 & 6.2153 & 6.5142 & 0.8748\\
					\textbf{SRD}~\cite{vicinanza2014pansharpening} & 0.8400 & 6.2484 & 6.3355 & 0.8732\\
					\textbf{PanNet}~\cite{yang2017pannet}& 0.9815 & 5.0513 & 3.6133 & 0.9676 \\ 
					\textbf{Fusion-Net}~\cite{deng2020detail} & \underline{0.9811}  & \underline{4.8672}  & \underline{3.4753}  & \underline{0.9777} \\
					\textbf{Our} & \textbf{0.9837}  & \textbf{4.7408}  & \textbf{3.3942}  & \textbf{0.9809} \\
					\midrule
					\textbf{Ideal value} & 1  & 0  & 0  & 1 \\
					\bottomrule		
				\end{tabular}	
			}
		}
	\end{center}	
\end{table}

\begin{figure*}[!htb]
	\centering
	\begin{minipage}{0.13\linewidth}
		\centerline{\includegraphics[width=1\linewidth]{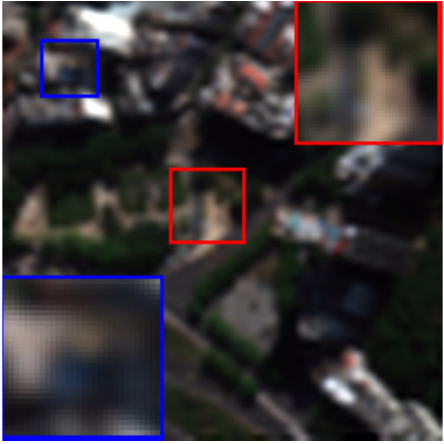}}
		\vspace{-1mm}
		\centerline{\footnotesize{(a) EXP}}
	\end{minipage}
	\begin{minipage}{0.13\linewidth}
		\centerline{\includegraphics[width=1\linewidth]{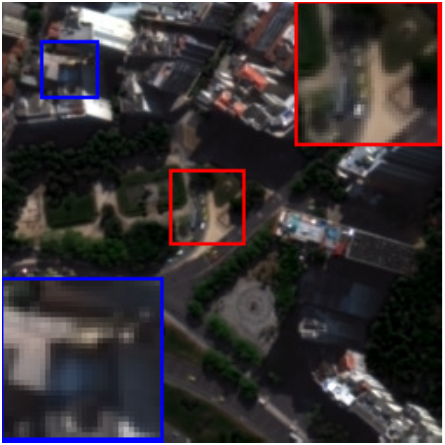}}
		\vspace{-1mm}
		\centerline{\footnotesize{(b) GS}}
	\end{minipage}
	\begin{minipage}{0.13\linewidth}
		\centerline{\includegraphics[width=1\linewidth]{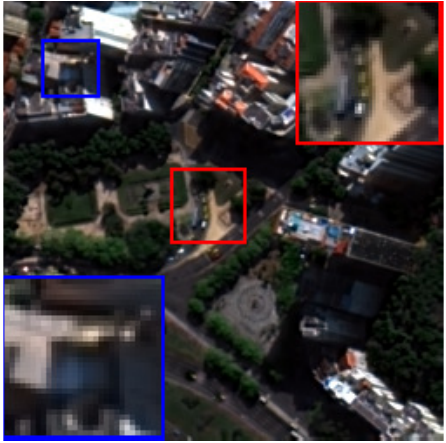}}
		\vspace{-1mm}
		\centerline{\footnotesize{(c) BDSD}}
	\end{minipage}
	\begin{minipage}{0.13\linewidth}
		\centerline{\includegraphics[width=1\linewidth]{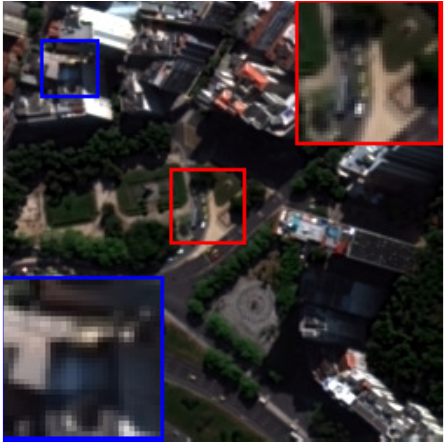}}
		\vspace{-1mm}
		\centerline{\footnotesize{(d) BDSD-PC}}
	\end{minipage}
	\begin{minipage}{0.13\linewidth}
		\centerline{\includegraphics[width=1\linewidth]{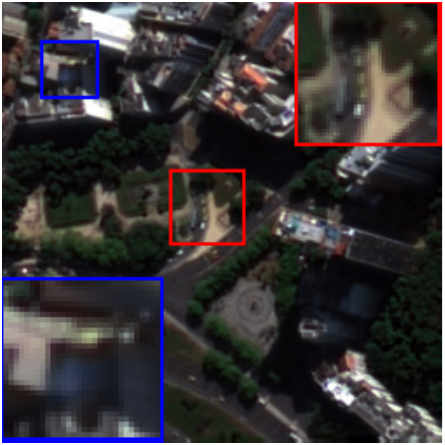}}
		\vspace{-1mm}
		\centerline{\footnotesize{(e) PRACS}}
	\end{minipage}
	\begin{minipage}{0.13\linewidth}
		\centerline{\includegraphics[width=1\linewidth]{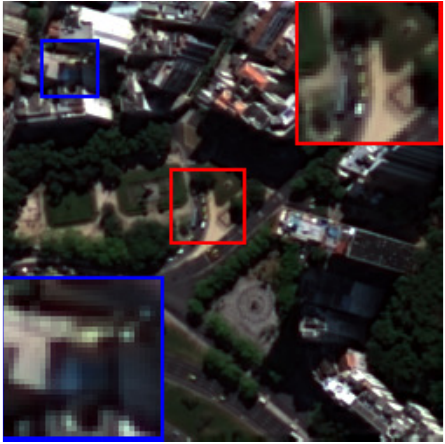}}
		\vspace{-1mm}
		\centerline{\footnotesize{(f) AWLP}}
	\end{minipage}
	\begin{minipage}{0.13\linewidth}
		\centerline{\includegraphics[width=1\linewidth]{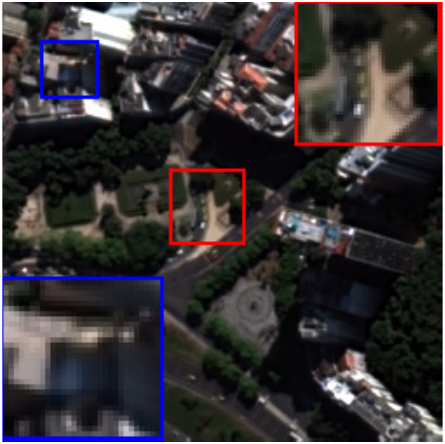}}
		\vspace{-1mm}
		\centerline{\footnotesize{(g) GLP}}
	\end{minipage}
	\vfill
	\begin{minipage}{0.13\linewidth}
		\centerline{\includegraphics[width=1\linewidth]{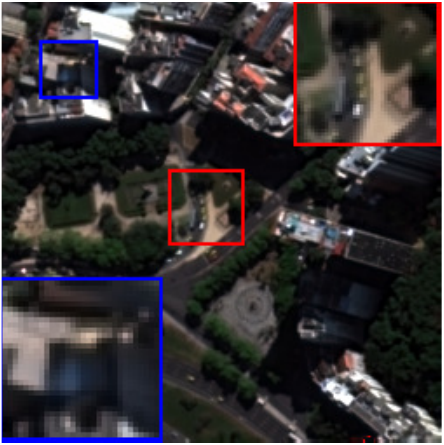}}
		\vspace{-1mm}
		\centerline{\footnotesize{(h) GLP-HPM}}
	\end{minipage}
	\begin{minipage}{0.13\linewidth}
		\centerline{\includegraphics[width=1\linewidth]{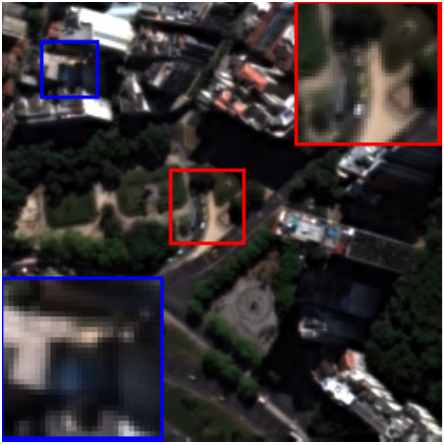}}
		\vspace{-1mm}
		\centerline{\footnotesize{(i) MO }}
	\end{minipage}
	\begin{minipage}{0.13\linewidth}
		\centerline{\includegraphics[width=1\linewidth]{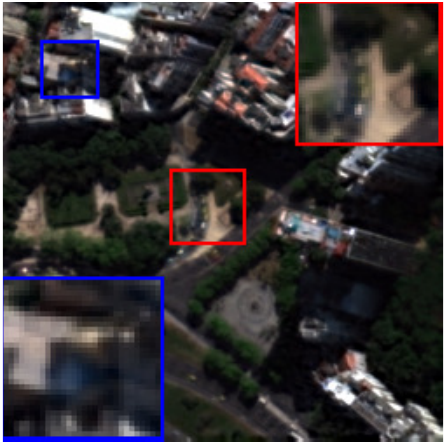}}
		\vspace{-1mm}
		\centerline{\footnotesize{(j) SRD}}
	\end{minipage}
	\begin{minipage}{0.13\linewidth}
		\centerline{\includegraphics[width=1\linewidth]{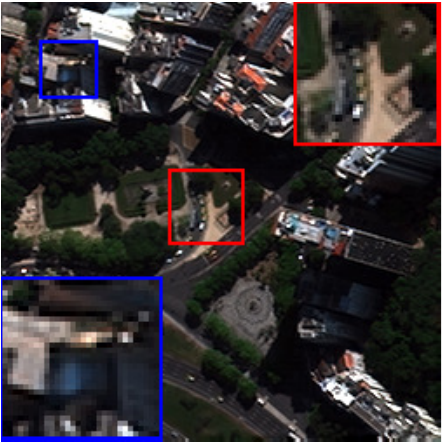}}
		\vspace{-1mm}
		\centerline{\footnotesize{(k) PanNet }}
	\end{minipage}
	\begin{minipage}{0.13\linewidth}
		\centerline{\includegraphics[width=1\linewidth]{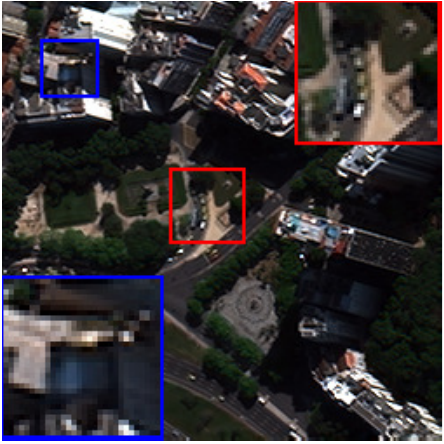}}
		\vspace{-1mm}
		\centerline{\footnotesize{(l) Fusion-Net}}
	\end{minipage}
	\begin{minipage}{0.13\linewidth}
		\centerline{\includegraphics[width=1\linewidth]{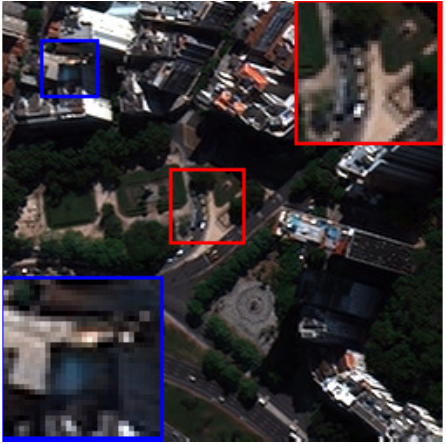}}
		\vspace{-1mm}
		\centerline{\footnotesize{(m) Our}}
	\end{minipage}	
	\begin{minipage}{0.13\linewidth}
		\centerline{\includegraphics[width=1\linewidth]{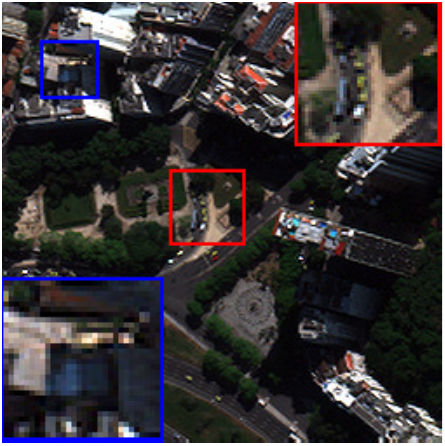}}
		\vspace{-1mm}
		\centerline{\footnotesize{(n) GT }}
	\end{minipage}
	\vspace{2mm}
	\caption{Visual comparisons of the fused HRMS image for all the methods on the Tripoli dataset.}
	\label{rr_Tripoli}
\end{figure*}

\begin{figure*}[!htb]
	\centering
	\begin{minipage}{0.13\linewidth}
		\centerline{\includegraphics[width=1\linewidth]{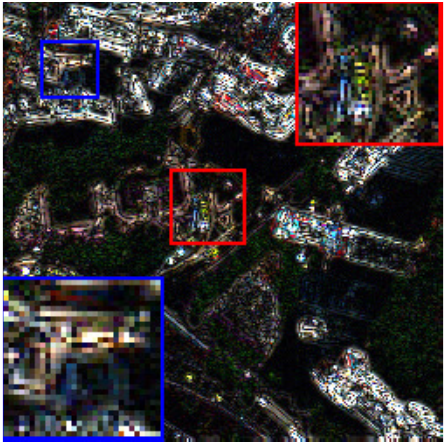}}
		\vspace{-1mm}
		\centerline{\footnotesize{(a) EXP}}
	\end{minipage}
	\begin{minipage}{0.13\linewidth}
		\centerline{\includegraphics[width=1\linewidth]{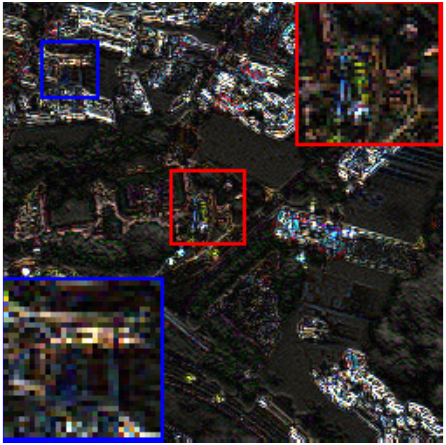}}
		\vspace{-1mm}
		\centerline{\footnotesize{(b) GS}}
	\end{minipage}
	\begin{minipage}{0.13\linewidth}
		\centerline{\includegraphics[width=1\linewidth]{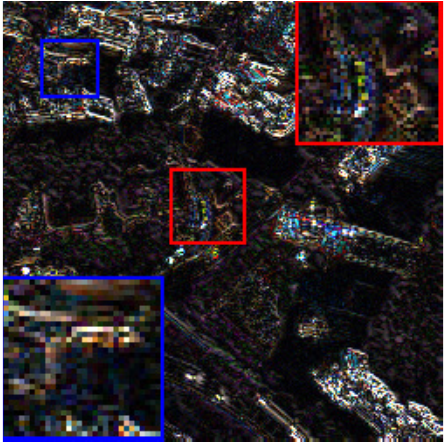}}
		\vspace{-1mm}
		\centerline{\footnotesize{(c) BDSD}}
	\end{minipage}
	\begin{minipage}{0.13\linewidth}
		\centerline{\includegraphics[width=1\linewidth]{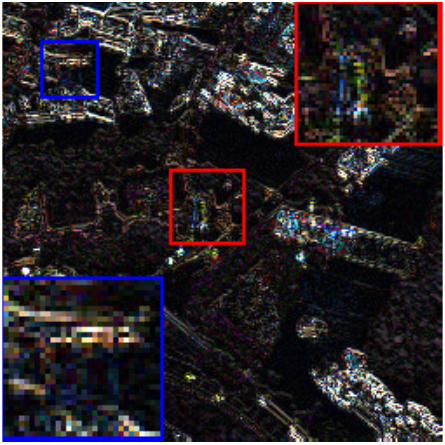}}
		\vspace{-1mm}
		\centerline{\footnotesize{(d) BDSD-PC}}
	\end{minipage}
	\begin{minipage}{0.13\linewidth}
		\centerline{\includegraphics[width=1\linewidth]{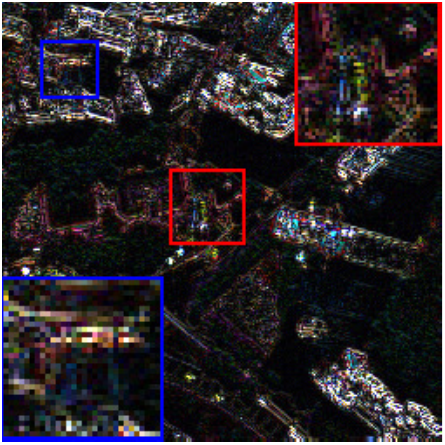}}
		\vspace{-1mm}
		\centerline{\footnotesize{(e) PRACS}}
	\end{minipage}
	\begin{minipage}{0.13\linewidth}
		\centerline{\includegraphics[width=1\linewidth]{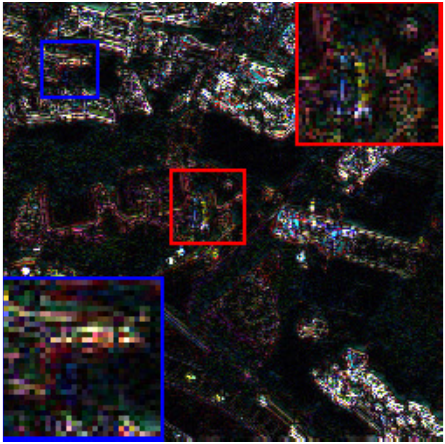}}
		\vspace{-1mm}
		\centerline{\footnotesize{(f) AWLP}}
	\end{minipage}
	\begin{minipage}{0.13\linewidth}
		\centerline{\includegraphics[width=1\linewidth]{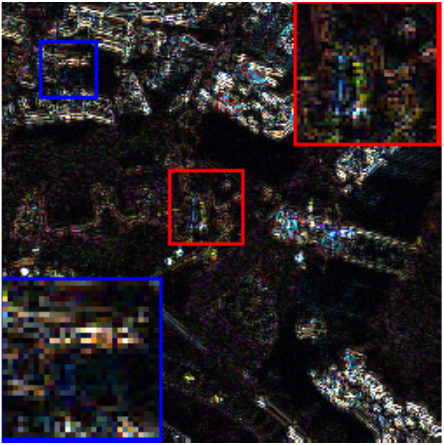}}
		\vspace{-1mm}
		\centerline{\footnotesize{(g) GLP}}
	\end{minipage}
	\vfill
	\begin{minipage}{0.13\linewidth}
		\centerline{\includegraphics[width=1\linewidth]{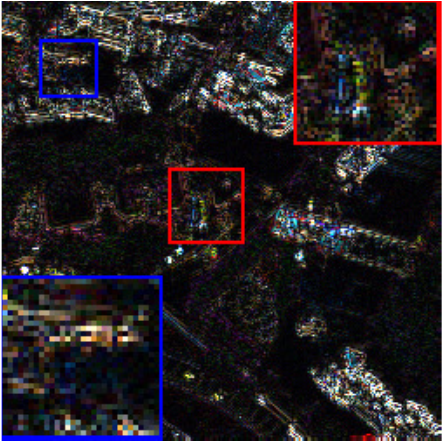}}
		\vspace{-1mm}
		\centerline{\footnotesize{(h) GLP-HPM}}
	\end{minipage}
	\begin{minipage}{0.13\linewidth}
		\centerline{\includegraphics[width=1\linewidth]{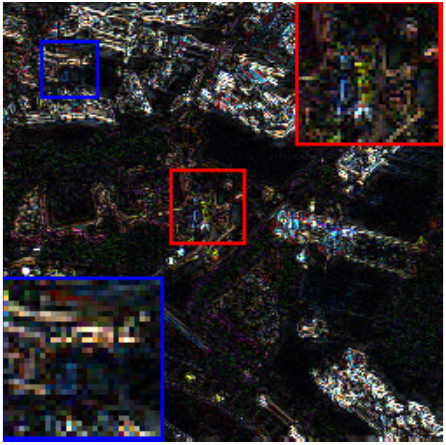}}
		\vspace{-1mm}
		\centerline{\footnotesize{(i) MO }}
	\end{minipage}
	\begin{minipage}{0.13\linewidth}
		\centerline{\includegraphics[width=1\linewidth]{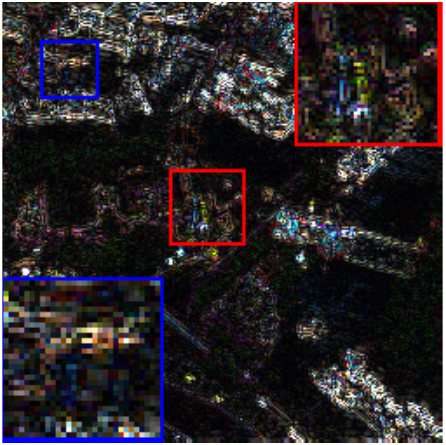}}
		\vspace{-1mm}
		\centerline{\footnotesize{(j) SRD}}
	\end{minipage}
	\begin{minipage}{0.13\linewidth}
		\centerline{\includegraphics[width=1\linewidth]{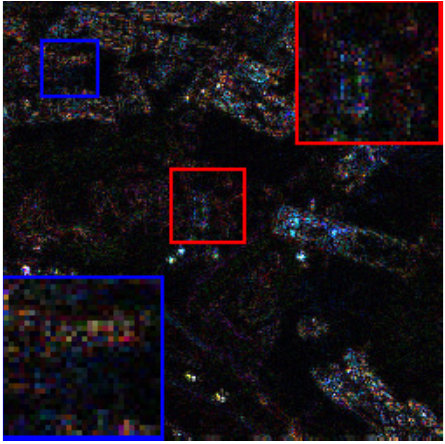}}
		\vspace{-1mm}
		\centerline{\footnotesize{(k) PanNet }}
	\end{minipage}
	\begin{minipage}{0.13\linewidth}
		\centerline{\includegraphics[width=1\linewidth]{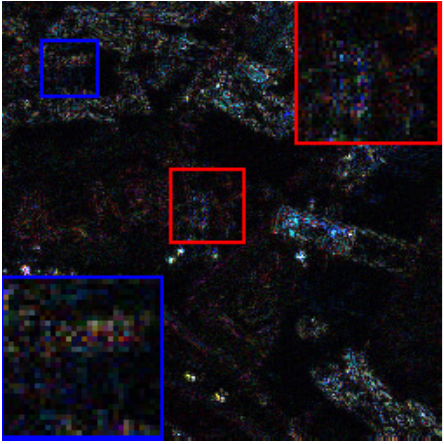}}
		\vspace{-1mm}
		\centerline{\footnotesize{(l) Fusion-Net}}
	\end{minipage}
	\begin{minipage}{0.13\linewidth}
		\centerline{\includegraphics[width=1\linewidth]{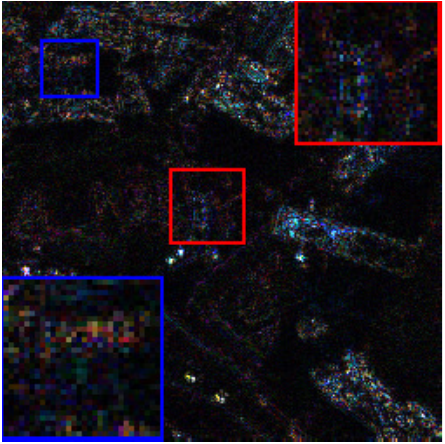}}
		\vspace{-1mm}
		\centerline{\footnotesize{(m) Our}}
	\end{minipage}	
	\begin{minipage}{0.13\linewidth}
		\centerline{\includegraphics[width=1\linewidth]{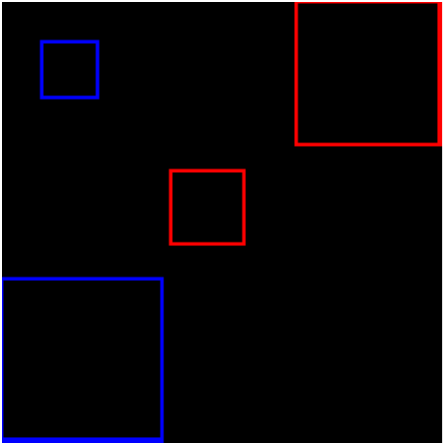}}
		\vspace{-1mm}
		\centerline{\footnotesize{(n) GT}}
	\end{minipage}	
	\vspace{2mm}
	\caption{The absolute value of residuals between the true HRMS and the fused HRMS of Figure~\ref{rr_Tripoli}.}
	\label{rr_Tripoli_res}
\end{figure*}

Further, we conduct another experiment on one new WorldView-3 dataset (i.e., Tripoli dataset), which has never been seen in the training phase. This dataset contains one PAN image (256$\times$256), one MS image (64$\times$64$\times$8) and one true HRMS image (256$\times$256$\times$8). The quantitative results are reported in Table~\ref{result_rr_1}, from which we can observe that our network achieves the best performance compared with other approaches. The visual comparison shown in Figure~\ref{rr_Tripoli} further verifies the numerical results. For better illustration, we enlarge two regions of the fused image. Specifically, it can be seen from Figure~\ref{rr_Tripoli} that the visual effect of classical methods contain obvious blur effects, while the three DL based methods (e.g., PanNet, FusionNet, and our network) produce better fused visual results. Additionally, we show the absolute value of residuals between the true HRMS and the predicted HRMS in Figure~\ref{rr_Tripoli_res}, from which we can observe that the residual image of our method contains fewer details.

\subsection{Evaluation at Full Resolution}
\begin{figure*}[!htb]
	\centering
	\begin{minipage}{0.13\linewidth}
		\centerline{\includegraphics[width=1\linewidth]{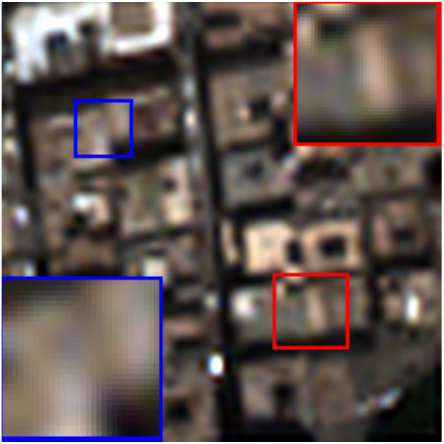}}
		\vspace{-1mm}
		\centerline{\footnotesize{(a) EXP}}
	\end{minipage}
	\begin{minipage}{0.13\linewidth}
		\centerline{\includegraphics[width=1\linewidth]{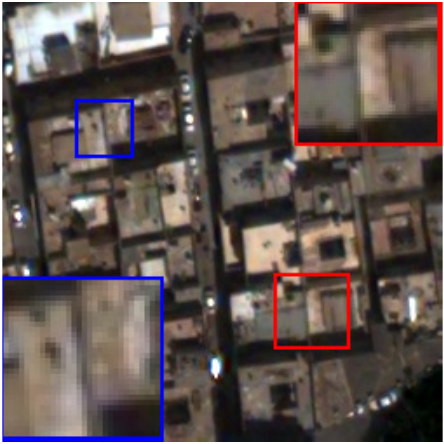}}
		\vspace{-1mm}
		\centerline{\footnotesize{(b) GS}}
	\end{minipage}
	\begin{minipage}{0.13\linewidth}
		\centerline{\includegraphics[width=1\linewidth]{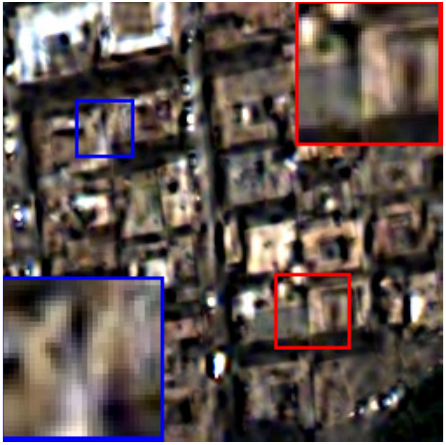}}
		\vspace{-1mm}
		\centerline{\footnotesize{(c) BDSD}}
	\end{minipage}
	\begin{minipage}{0.13\linewidth}
		\centerline{\includegraphics[width=1\linewidth]{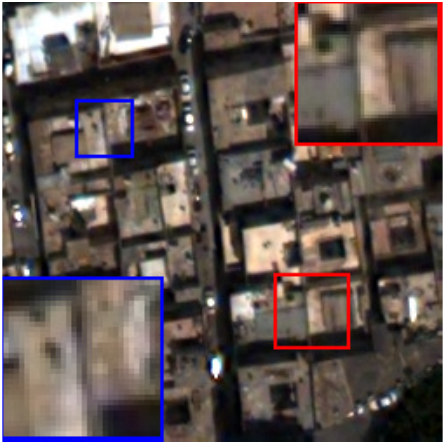}}
		\vspace{-1mm}
		\centerline{\footnotesize{(d) BDSD-PC}}
	\end{minipage}
	\begin{minipage}{0.13\linewidth}
		\centerline{\includegraphics[width=1\linewidth]{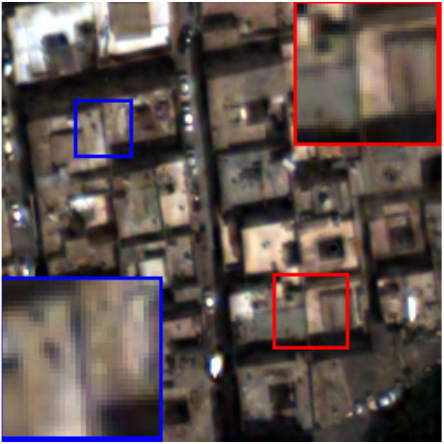}}
		\vspace{-1mm}
		\centerline{\footnotesize{(e) PRACS}}
	\end{minipage}
	\begin{minipage}{0.13\linewidth}
		\centerline{\includegraphics[width=1\linewidth]{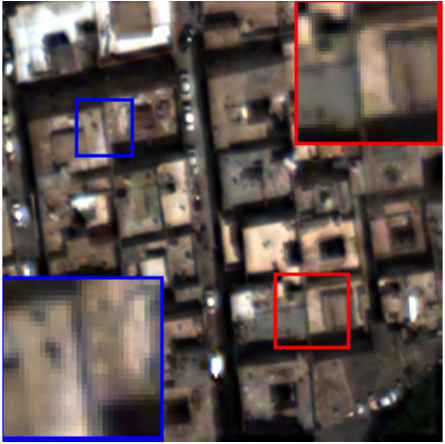}}
		\vspace{-1mm}
		\centerline{\footnotesize{(f) AWLP}}
	\end{minipage}
	\begin{minipage}{0.13\linewidth}
		\centerline{\includegraphics[width=1\linewidth]{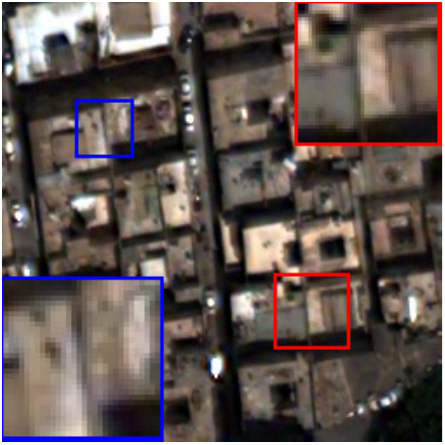}}
		\vspace{-1mm}
		\centerline{\footnotesize{(g) GLP}}
	\end{minipage}
	\vfill
	\begin{minipage}{0.13\linewidth}
		\centerline{\includegraphics[width=1\linewidth]{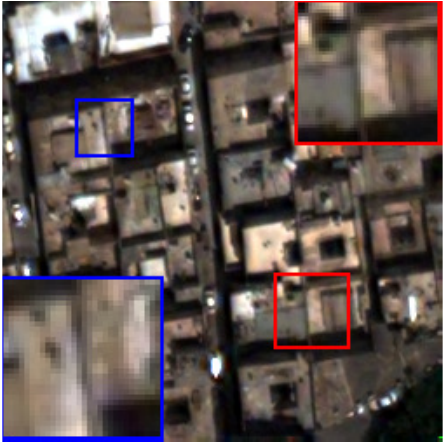}}
		\vspace{-1mm}
		\centerline{\footnotesize{(h) GLP-HPM}}
	\end{minipage}
	\begin{minipage}{0.13\linewidth}
		\centerline{\includegraphics[width=1\linewidth]{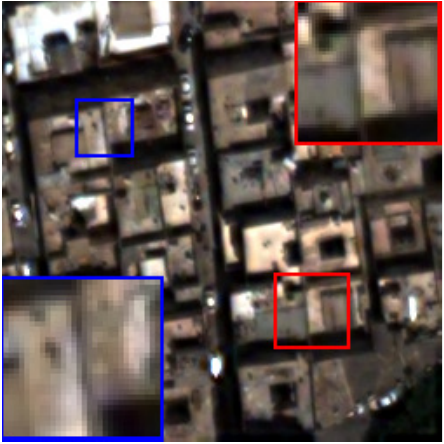}}
		\vspace{-1mm}
		\centerline{\footnotesize{(i) MO }}
	\end{minipage}
	\begin{minipage}{0.13\linewidth}
		\centerline{\includegraphics[width=1\linewidth]{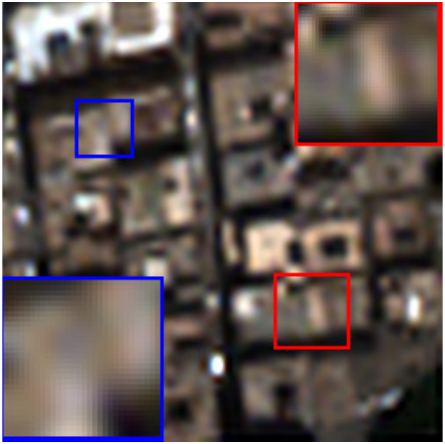}}
		\vspace{-1mm}
		\centerline{\footnotesize{(j) SRD}}
	\end{minipage}
	\begin{minipage}{0.13\linewidth}
		\centerline{\includegraphics[width=1\linewidth]{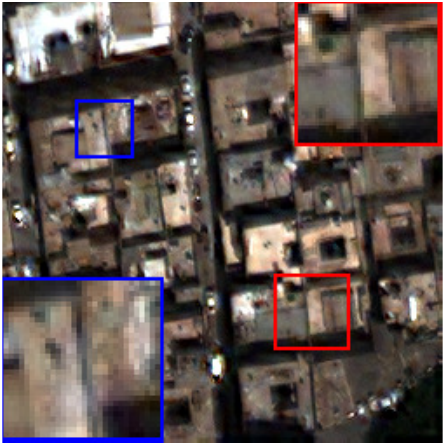}}
		\vspace{-1mm}
		\centerline{\footnotesize{(k) PanNet }}
	\end{minipage}
	\begin{minipage}{0.13\linewidth}
		\centerline{\includegraphics[width=1\linewidth]{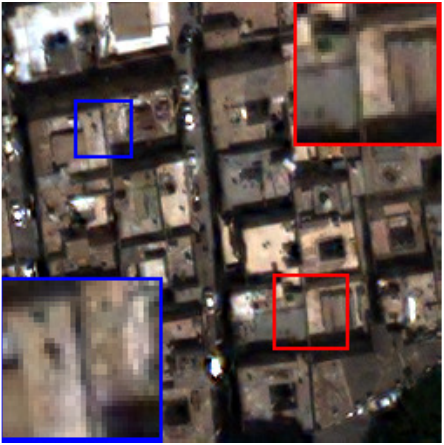}}
		\vspace{-1mm}
		\centerline{\footnotesize{(l) Fusion-Net}}
	\end{minipage}
	\begin{minipage}{0.13\linewidth}
		\centerline{\includegraphics[width=1\linewidth]{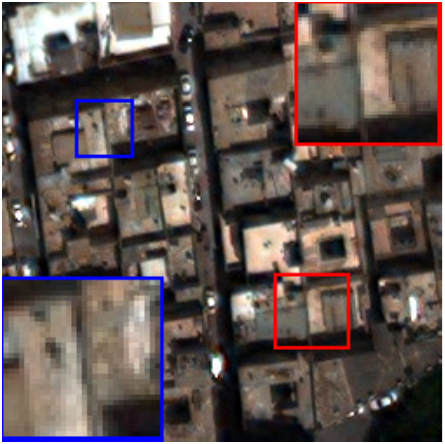}}
		\vspace{-1mm}
		\centerline{\footnotesize{(m) Our}}
	\end{minipage}	
	\vspace{2mm}
	\caption{Visual comparisons of the fused HRMS image for all the methods on a full resolution sample.}
	\label{fr_sample}
\end{figure*}

This section evaluates the performance of our network in the full resolution case. We conduct this experiment on 30 image pairs from the Worldview3 satellite. Each image pair contains one PAN image (256$\times$256), and one MS image (64$\times$64$\times$8). Since the ground-truth HRMS image is not available, we thus adopt $D_{\lambda}$, $D_{s}$, and QNR~\cite{vivone2014critical} as the indexes. The quantitative results are reported in Table~\ref{result_or}. As can be seen, the DL-based methods (i.e., PanNet, Fusion-Net, and our network) perform better than other methods, which verifies the superiority of the DL methods. Among the three DL approaches, our network performs best. Additionally, we also show visual comparison of a full-resolution sample in Figure~\ref{fr_sample}, which indicates that our network obtains the best visual effect both spatially and spectrally than the classical approaches, and also has a comparable visual effect with the other two DL methods.

\begin{table}
	\setlength{\abovecaptionskip}{0cm}
	\setlength{\belowcaptionskip}{-0.15cm}
	\caption{\label{result_or} Average quantitative results on the WorldView-3 dataset in the full resolution case (30 samples).}
	\begin{center}
		{\small
			\scalebox{0.6}[0.6]{			
				\begin{tabular}{l|c c c }
					\toprule
					\textbf{Methods} &\textbf{$D_{\lambda}$}&\textbf{$D_{s}$}&\textbf{QNR}\\
					\midrule
					\textbf{EXP}~\cite{aiazzi2002context} & 0.0104$\pm$0.0035 & 0.0594$\pm$0.0137 & 0.9276$\pm$0.0125 \\
					\textbf{GS}~\cite{laben2000process} & 0.0275$\pm$0.0072 & 0.0686$\pm$0.0190 & 0.9058$\pm$0.0221 \\
					\textbf{BDSD}~\cite{garzelli2007optimal} & 0.0094$\pm$0.0020 & 0.0429$\pm$0.0073 & 0.9482$\pm$0.0083  \\
					\textbf{BDSD-PC}~\cite{vivone2019robust} & \underline{0.0092$\pm$0.0026} & 0.0434$\pm$0.0078 & 0.9478$\pm$0.0095 \\
					\textbf{PRACS}~\cite{choi2010new} & 0.0109$\pm$0.0025 & 0.0365$\pm$0.0071 & 0.9528$\pm$0.0081 \\
					\textbf{AWLP}~\cite{otazu2005introduction} & 0.0161$\pm$0.0328 & 0.0325$\pm$0.0056 & 0.9518$\pm$0.0067 \\
					\textbf{GLP}~\cite{restaino2019pansharpening} & 0.0176$\pm$0.0062 & 0.0385$\pm$0.0067 & 0.9445$\pm$0.0117 \\
					\textbf{GLP-HPM}~\cite{vivone2013contrast} & 0.0157$\pm$0.0055 & 0.0365$\pm$0.0063 & 0.9485$\pm$0.0106 \\
					\textbf{MO}~\cite{restaino2016fusion} & 0.0219$\pm$0.0069 & 0.0356$\pm$0.0059 & 0.9433$\pm$0.0112  \\
					\textbf{SRD}~\cite{vicinanza2014pansharpening} & 0.0097$\pm$0.0026 & 0.0614$\pm$0.0160 & 0.9345$\pm$0.0184  \\
					\textbf{PanNet}~\cite{yang2017pannet}& 0.0094$\pm$0.0026 & 0.0328$\pm$0.0044 & 0.9541$\pm$0.0065  \\ 
					\textbf{Fusion-Net}~\cite{deng2020detail} & 0.0093$\pm$0.0024  & \underline{0.0320$\pm$0.0047}  & \underline{0.9560$\pm$0.0053}  \\
					\textbf{Our} & \textbf{0.0086$\pm$0.0028}  & \textbf{0.0315$\pm$0.0046}  & \textbf{0.9606$\pm$0.0059} \\
					\midrule
					\textbf{Ideal value} & 0  & 0  & 1 \\
					\bottomrule		
				\end{tabular}	
			}
		}
	\end{center}	
\end{table}

\subsection{Generalization to New Satellites}
\begin{figure*}[!htb]
	\centering	
	\begin{minipage}{0.13\linewidth}
		\centerline{\includegraphics[width=1\linewidth]{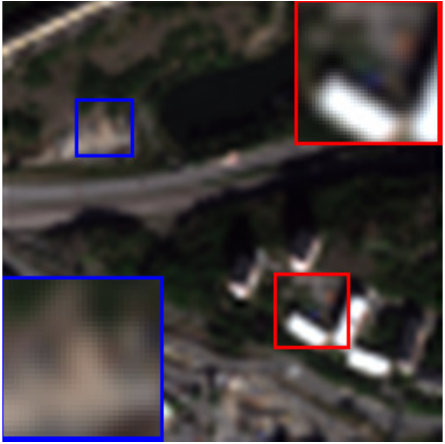}}
		\vspace{-1mm}
		\centerline{\footnotesize{(a) EXP}}
	\end{minipage}
	\begin{minipage}{0.13\linewidth}
		\centerline{\includegraphics[width=1\linewidth]{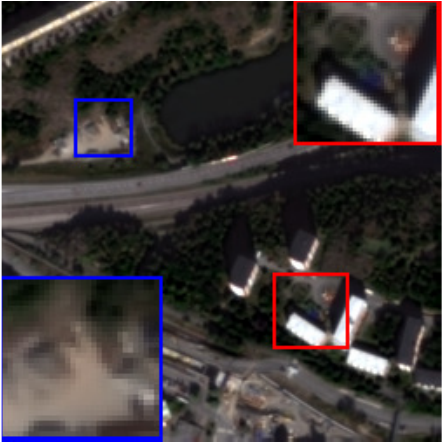}}
		\vspace{-1mm}
		\centerline{\footnotesize{(b) GS}}
	\end{minipage}
	\begin{minipage}{0.13\linewidth}
		\centerline{\includegraphics[width=1\linewidth]{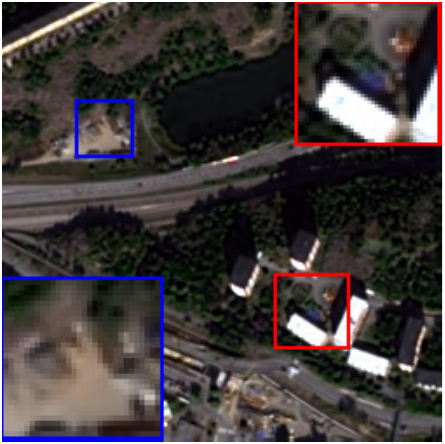}}
		\vspace{-1mm}
		\centerline{\footnotesize{(c) BDSD}}
	\end{minipage}
	\begin{minipage}{0.13\linewidth}
		\centerline{\includegraphics[width=1\linewidth]{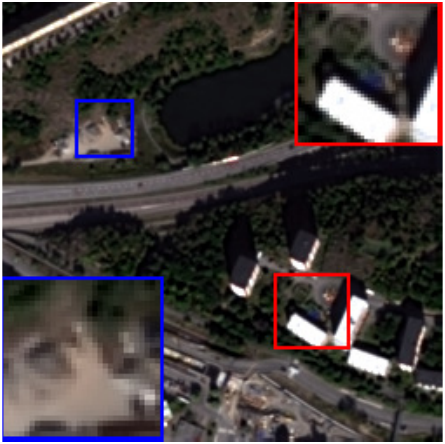}}
		\vspace{-1mm}
		\centerline{\footnotesize{(d) BDSD-PC}}
	\end{minipage}
	\begin{minipage}{0.13\linewidth}
		\centerline{\includegraphics[width=1\linewidth]{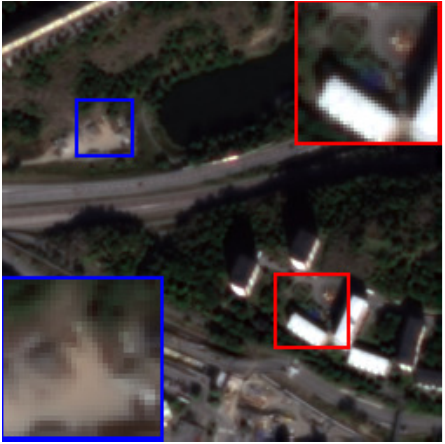}}
		\vspace{-1mm}
		\centerline{\footnotesize{(e) PRACS}}
	\end{minipage}
	\begin{minipage}{0.13\linewidth}
		\centerline{\includegraphics[width=1\linewidth]{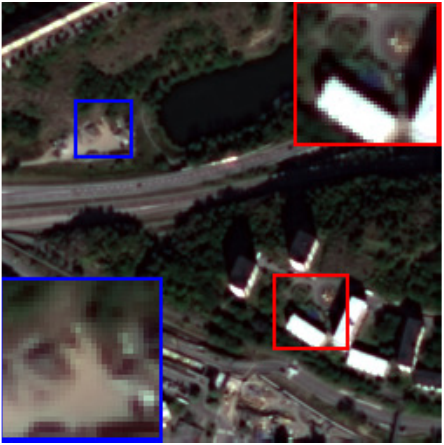}}
		\vspace{-1mm}
		\centerline{\footnotesize{(f) AWLP}}
	\end{minipage}
	\begin{minipage}{0.13\linewidth}
		\centerline{\includegraphics[width=1\linewidth]{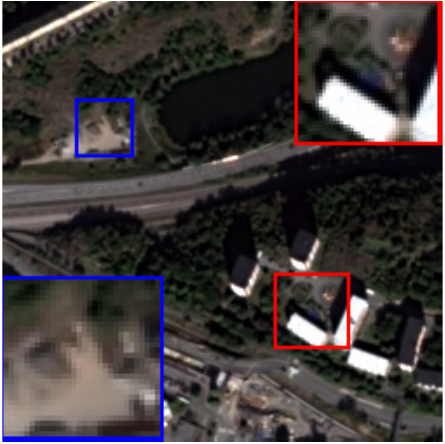}}
		\vspace{-1mm}
		\centerline{\footnotesize{(g) GLP}}
	\end{minipage}
	\vfill
	\begin{minipage}{0.13\linewidth}
		\centerline{\includegraphics[width=1\linewidth]{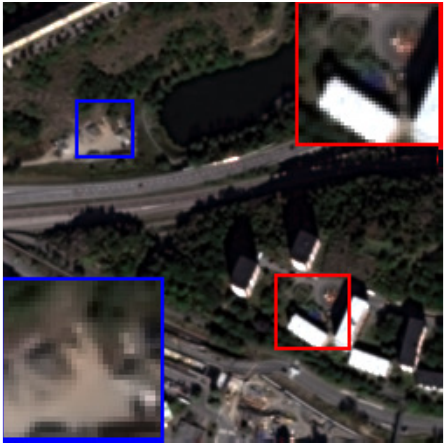}}
		\vspace{-1mm}
		\centerline{\footnotesize{(h) GLP-HPM}}
	\end{minipage}
	\begin{minipage}{0.13\linewidth}
		\centerline{\includegraphics[width=1\linewidth]{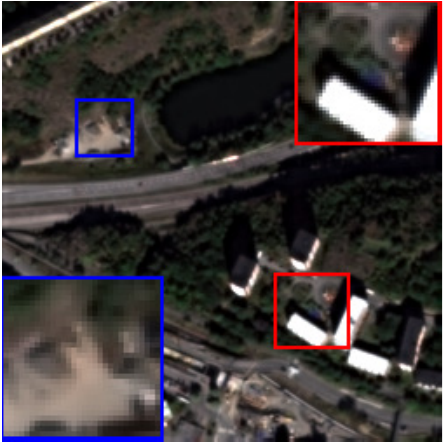}}
		\vspace{-1mm}
		\centerline{\footnotesize{(i) MO }}
	\end{minipage}
	\begin{minipage}{0.13\linewidth}
		\centerline{\includegraphics[width=1\linewidth]{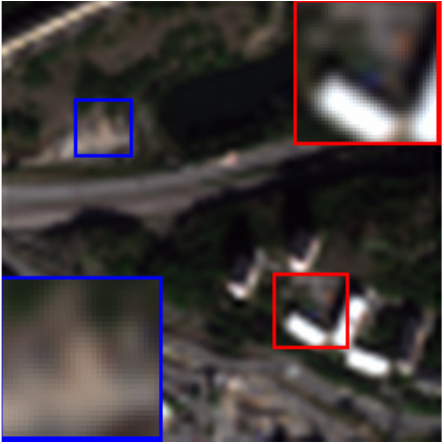}}
		\vspace{-1mm}
		\centerline{\footnotesize{(j) SRD}}
	\end{minipage}
	\begin{minipage}{0.13\linewidth}
		\centerline{\includegraphics[width=1\linewidth]{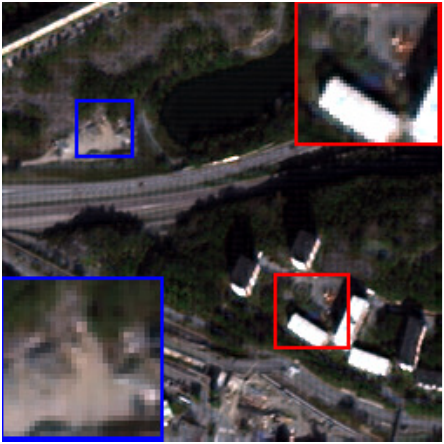}}
		\vspace{-1mm}
		\centerline{\footnotesize{(k) PanNet }}
	\end{minipage}
	\begin{minipage}{0.13\linewidth}
		\centerline{\includegraphics[width=1\linewidth]{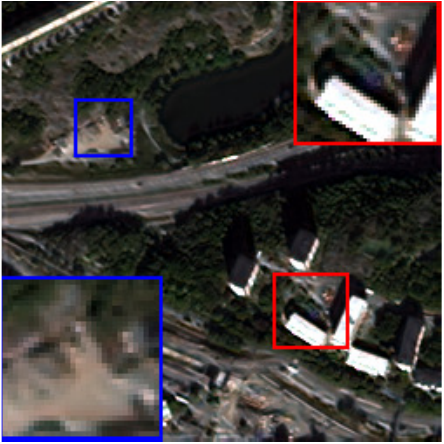}}
		\vspace{-1mm}
		\centerline{\footnotesize{(l) Fusion-Net}}
	\end{minipage}
	\begin{minipage}{0.13\linewidth}
		\centerline{\includegraphics[width=1\linewidth]{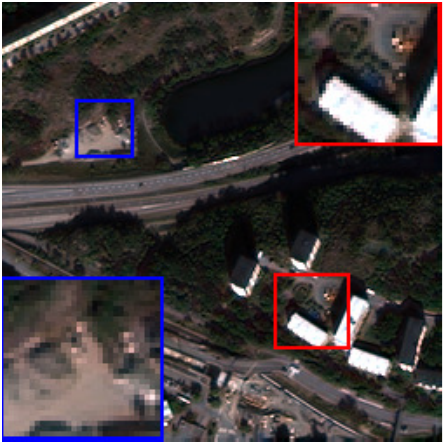}}
		\vspace{-1mm}
		\centerline{\footnotesize{(m) Our}}
	\end{minipage}	
	\begin{minipage}{0.13\linewidth}
		\centerline{\includegraphics[width=1\linewidth]{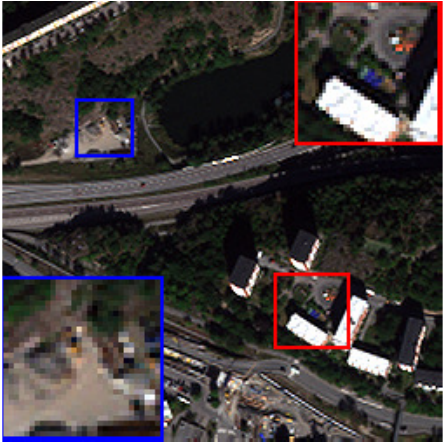}}
		\vspace{-1mm}
		\centerline{\footnotesize{(n) GT}}
	\end{minipage}
	\vspace{2mm}
	\caption{Visual comparisons of the fused HRMS image for all the methods on the WorldView-2 dataset.}
	\label{gener_new}
\end{figure*}

In this section, we explore the generalization of our network on the images acquired by other satellites. Since our network is trained using the WolrdView-3 images, we thus test our network on the images generated by the WorldView-2 satellite. We also adopt the Wald’s protocol to generate 4 test image pairs from the WorldView-2 image. Each image pair contains one PAN image (256$\times$256), one MS image (64$\times$64$\times$8), and one true HRMS image (256$\times$256$\times$8). The average quantitative results are reported in Table~\ref{result_gener}. As can be seen, our network has the best performance concerning the SAM, ERGAS, and SCC, and obtains the second-best performance in terms of the Q8 index. These results indicate that our network performs almost the best in both spatial and spectral domain, which implies that our network can generalize well to new satelites. Additionally, we also illustrate the visual results of all the competing methods in Figure~\ref{gener_new}. For easy observation, two small areas of this image are amplified. Figure~\ref{gener_new} shows that our network has clearer visual results, while other methods contain obvious blur effects, which further verifies the superiority of our network.

\begin{table}
	\setlength{\abovecaptionskip}{0cm}
	\setlength{\belowcaptionskip}{-0.2cm}
	\caption{\label{result_gener} Average quantitative result on WorldView-2 dataset (4 samples).}
	\begin{center}
		{\small
			\scalebox{0.53}[0.53]{			
				\begin{tabular}{l|c c c c }
					\toprule
					\textbf{Methods} &\textbf{Q8}&\textbf{SAM}&\textbf{ERGAS}&\textbf{SCC}\\
					\midrule
					\textbf{EXP}~\cite{aiazzi2002context}  & 0.1478$\pm$0.0993  &8.4696$\pm$0.6908 & 8.4482$\pm$0.8586 & 0.7305$\pm$0.0291\\
					\textbf{GS}~\cite{laben2000process} & 0.3699$\pm$0.0973 & 8.9007$\pm$1.1376 & 6.0507$\pm$0.3289 & 0.8982$\pm$0.0157\\
					\textbf{BDSD}~\cite{garzelli2007optimal} & 0.3813$\pm$0.0792 & 8.1413$\pm$0.8517 & 5.6247$\pm$0.6358 & 0.8986$\pm$0.0154\\
					\textbf{BDSD-PC}~\cite{vivone2019robust} & 0.2260$\pm$0.0792 & 8.5559$\pm$0.8517 & 6.7658$\pm$0.6358 & 0.8727$\pm$0.0154\\
					\textbf{PRACS}~\cite{choi2010new} & 0.2344$\pm$0.1275 & 8.2384$\pm$0.8070 & 6.7702$\pm$0.5389 & 0.8552$\pm$0.0317\\
					\textbf{AWLP}~\cite{otazu2005introduction} & \textbf{0.4048$\pm$0.1339} & 8.0334$\pm$0.7286 & 5.8288$\pm$0.6002 & 0.8947$\pm$0.0318\\
					\textbf{GLP}~\cite{restaino2019pansharpening} & 0.3777$\pm$0.0653 & 7.9013$\pm$0.8200 & 5.5841$\pm$0.5361 & 0.8942$\pm$0.0216\\
					\textbf{GLP-HPM}~\cite{vivone2013contrast} & 0.3171$\pm$0.0741 & 8.0942$\pm$0.9487 & 5.8434$\pm$0.6011 & 0.8826$\pm$0.0220\\
					\textbf{MO}~\cite{restaino2016fusion} & 0.3381$\pm$0.1108 & 7.8839$\pm$0.9794 & 5.9212$\pm$0.5137 & 0.8936$\pm$0.0102\\
					\textbf{SRD}~\cite{vicinanza2014pansharpening} & 0.1478$\pm$0.0993 & 8.4696$\pm$0.6908 & 8.4482$\pm$0.8580 & 0.7305$\pm$0.0291\\
					\textbf{PanNet}~\cite{yang2017pannet}& 0.3746$\pm$0.1053 & \underline{7.5776$\pm$0.6670} & \underline{5.5402$\pm$0.1744} & \underline{0.9144$\pm$0.0178} \\ 
					\textbf{Fusion-Net}~\cite{deng2020detail} & 0.3685$\pm$0.0797 & 8.2145$\pm$1.1321 & 6.1401$\pm$0.3592 & 0.8890$\pm$0.0187 \\
					\textbf{Our} & \underline{0.3885$\pm$0.1337}  & \textbf{7.5233$\pm$0.9893} & \textbf{5.4103$\pm$0.1841}  & \textbf{0.9231$\pm$0.0120} \\
					\midrule
					\textbf{Ideal value} & 1  & 0  & 0  & 1 \\
					\bottomrule		
				\end{tabular}	
			}
		}
	\end{center}	
\end{table}

\subsection{Parameters and Time Comparisons} In Table~\ref{testing_time},  we also report the running time\footnote{For the DL approaches, we only record the test time.}, the number of parameters, and the giga floating-point operations per second (GFLOPs) of our network with some typical methods on two image sizes. From Table~\ref{testing_time}, we can find that our network has a comparable efficiency with the advanced approaches. Besides, our network has fewer parameters compared to the other two DL-based methods.

\begin{table}
	\setlength{\abovecaptionskip}{0cm}
	\setlength{\belowcaptionskip}{-0.2cm}
	\caption{\label{testing_time} Parameters and time comparisons.}
	\begin{center}
		{\small
			\scalebox{0.54}[0.54]{			
				\begin{tabular}{c|c|c|c|c|c|c|c}
					\toprule
					& \textbf{Image size}& \textbf{BDSD} & \textbf{AWLP} &  \textbf{GLP}  & \textbf{PanNet} & \textbf{Fusion-Net} & \textbf{Our}\\
					\midrule
					\multirow{2}{*}{Runtime (s)}
					& \tabincell{c}{64$\times$64 (PAN) \\16$\times$16$\times$8 (MS)}&  \textbf{0.01}  & \textbf{0.01} & 0.03  & 0.42 & 0.44 & 0.65\\
					\cline{2-8}
					& \tabincell{c}{256$\times$256 (PAN) \\ 64$\times$64$\times$8 (MS)}  & \textbf{0.02}  & 0.08 & 0.12 & 2.63 & 2.31 & 2.79 \\
					\hline
					\multirow{2}{*}{GFLOPs}
					& \tabincell{c}{64$\times$64 (PAN) \\16$\times$16$\times$8 (MS)} & \textbf{ $4.2\times 10^{-8}$}   & $5.9\times 10^{-4}$ & $6.5\times 10^{-5}$  &  0.65 & 0.64 & 0.71\\
					\cline{2-8}
					& \tabincell{c}{256$\times$256 (PAN) \\ 64$\times$64$\times$8 (MS)}   & \textbf{ $4.2\times 10^{-8}$ }  & $5.9\times 10^{-4}$ & $6.5\times 10^{-5}$  & 10.3 & 10.2 & 11.2 \\
					\hline
					$\#$ Params ($\times 10^4$) & -  & - & - & - & 7.76 & 7.63 & \textbf{7.03}\\
					\bottomrule		
				\end{tabular}	
			}
		}
	\end{center}	
\end{table}

\section{Conclusion}
This paper proposes a novel deep network for pansharpening using the model-based deep learning method. Specifically, we first propose a new pansharpening model based on the convolutional sparse coding technique. Then, we design a proximal gradient algorithm to solve this model, and then unfold this iterative algorithm into a network. Finally, all the learnable modules can be learned end-to-end. Since each network module corresponds to an operation of the algorithm, our network has a clear physical interpretation. This helps us to better understand how the network works. Experimental results on some benchmark datasets illustrate our network has better performance than other advanced approaches both quantitatively and qualitatively.

\bibliography{aaai22}
\end{document}